\documentclass{article}

\usepackage{microtype}
\usepackage{graphicx}
\usepackage{subfigure}
\usepackage{booktabs} % for professional tables
\usepackage{soul}
\usepackage{multirow}
\usepackage{array}
\usepackage{svg}
\usepackage{siunitx}
\usepackage{xspace}

\usepackage{hyperref}

\usepackage[accepted]{arxiv}

\usepackage{amsmath}
\usepackage{amssymb}
\usepackage{mathtools}
\usepackage{amsthm}

\usepackage[capitalize,noabbrev]{cleveref}

\usepackage{MnSymbol}

\theoremstyle{plain}

\theoremstyle{definition}

\theoremstyle{remark}

\usepackage[textsize=tiny]{todonotes}

\newcommand{\aPAC}{$\alpha$-\texttt{PAC}\xspace}
\newcommand{\PACQ}{\texttt{PAC}\xspace}
\newcommand{\PACV}{\texttt{PAC+V}\xspace}
\newcommand{\BC}{\texttt{BC+Q}\xspace}
\newcommand{\fBC}{\texttt{FilteredBC}\xspace}

\usepackage{amsmath}
\usepackage{amssymb}
\usepackage{nicefrac}
\usepackage{mathtools}
\usepackage{multirow}

\DeclareMathOperator*{\EXP}{\mathbb{E}}% expected value
\DeclareMathOperator{\KL}{D_{KL}}
\newcommand{\pit}{\tilde{\pi}}
\newcommand{\piimp}{\pi_{\text{imp}}}
\newcommand{\pipac}{\pi_{\theta}}
\newcommand{\task}{\tau}

\icmltitlerunning{Offline Actor-Critic Reinforcement Learning Scales to Large Models}

\begin{document}

\twocolumn[
\icmltitle{Offline Actor-Critic Reinforcement Learning Scales to Large Models}



\icmlsetsymbol{equal}{*}

\begin{icmlauthorlist}
\icmlauthor{Jost Tobias Springenberg}{equal,comp}
\icmlauthor{Abbas Abdolmaleki}{equal,comp}
\icmlauthor{Jingwei Zhang}{equal,comp}
\icmlauthor{Oliver Groth}{equal,comp}
\icmlauthor{Michael Bloesch}{equal,comp}
\icmlauthor{Thomas Lampe}{equal,comp}
\icmlauthor{Philemon Brakel}{equal,comp}
\icmlauthor{Sarah Bechtle}{equal,comp}
\icmlauthor{Steven Kapturowski}{equal,comp}
\icmlauthor{Roland Hafner}{equal,comp}
\icmlauthor{Nicolas Heess}{comp}
\icmlauthor{Martin Riedmiller}{comp}
\end{icmlauthorlist}

%\icmlaffiliation{yyy}{Department of XXX, University of YYY, Location, Country}
\icmlaffiliation{comp}{Google Deepmind, London, United Kingdom}
%\icmlaffiliation{sch}{School of ZZZ, Institute of WWW, Location, Country}

\icmlcorrespondingauthor{Jost Tobias Springenberg}{springenberg@google.com}
%\icmlcorrespondingauthor{Firstname2 Lastname2}{first2.last2@www.uk}

\icmlkeywords{Machine Learning, ICML}

\vskip 0.3in
]



\printAffiliationsAndNotice{\icmlEqualContribution} % otherwise use the standard text.

\begin{abstract}
 We show that offline actor-critic reinforcement learning can scale to large models -- such as transformers -- and follows similar scaling laws as supervised learning.
 We find that offline actor-critic algorithms can outperform strong, supervised, behavioral cloning baselines for multi-task training on a large dataset containing both sub-optimal and expert behavior on 132 continuous control tasks.
 We introduce a Perceiver-based actor-critic model and elucidate the key model features needed to make offline RL work with self- and cross-attention modules.
 Overall, we find that: i) simple offline actor critic algorithms are a natural choice for gradually moving away from the currently predominant paradigm of behavioral cloning, and ii) via offline RL it is possible to learn multi-task policies that master many domains simultaneously, including real robotics tasks, from sub-optimal demonstrations or self-generated data.
\end{abstract}

\section{Introduction}
In recent years, scaling both model and dataset sizes has led to multiple breakthroughs in machine learning.
In particular, generative pre-training of large (vision-)language models on diverse, web-scale data is now the standard way to solve many language and vision tasks \citep{openai2023gpt4,alayrac2022flamingo} and generative models of images and music have, in the last years, reached unprecedented quality \citep{rombach2021stable,kang2023gigagan}. 

Recent work on scaling up policy learning for control has shown that, when similar model architectures are used (e.g.\ transformers), supervised behaviour cloning (BC) from large datasets can lead to surprisingly capable multi-task policies \citep{reed2022generalist,bousmalis2023robocat,brohan2022rt,team2023octo}.
Although impressive in their capabilities, these examples come with the drawback that high-quality ('expert' demonstration) data is needed for training.
While such high quality data is readily available for language and vision domains via the internet, in robotics and other real world control domains expert data is at best scarce and expensive to obtain -- and in many cases it is not available in the first place.
It is thus desirable to use different training methods, such as reinforcement learning (RL), that can utilize sub-optimal data or data generated without a human in the loop, i.e.\ generated by an agent, -- which can be more readily available -- while retaining model architectures and scaling benefits. 

However, training large behaviour models via offline RL methods\footnote{This is in contrast to online RL of transformer models which is often applied when large language models are fine-tuned with RLHF, but is prohibitively expensive in real-world settings.} is a largely unexplored area of research.
While first explorations of applying pure Q-learning on larger multi-task datasets exist \citep{kumar2022offline,chebotar2023q} they either consider non-transformer models of moderate size \citep{kumar2022offline} or adapt relatively small models and incur significant computational overhead during training \citep{chebotar2023q}.
What is missing is a clear recipe detailing how to scale offline RL to large transformers accompanied by an efficient model.

\begin{figure*}[ht]
    \centering
    \includegraphics[width=.99\textwidth]{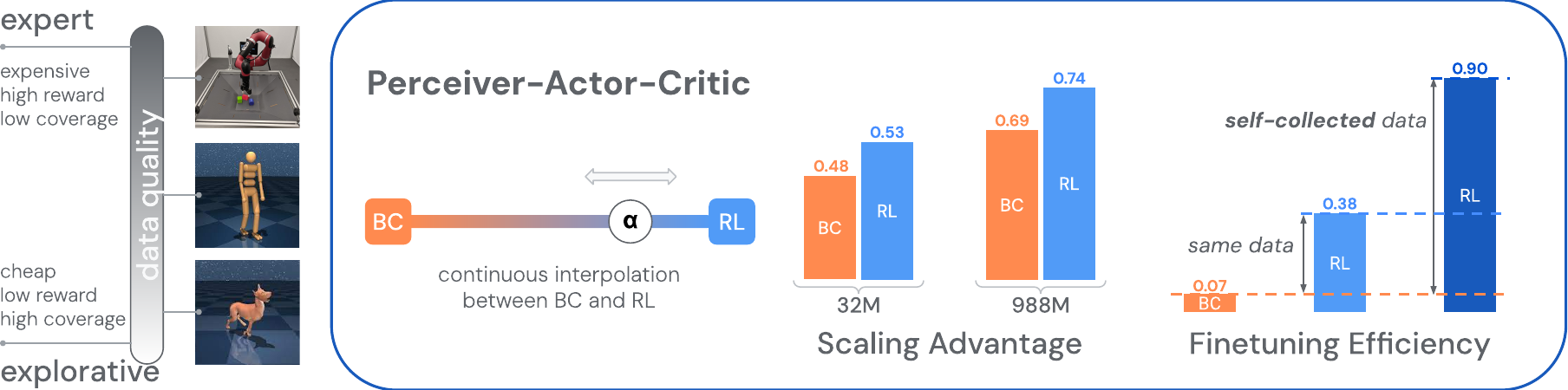}
    \caption{
    PAC is a scalable neural architecture for continuous control able to smoothly interpolate between BC and offline RL.
    The system design enables training on heterogenous, multi-modal data of varying quality.
    We demonstrate that our system achieves higher performance than BC across a series of model scales. The method also enables a seamless transition into offline and online RL finetuning for fast adaptation and mastery of control tasks.
    }
    \label{fig:pac-teaser}
\end{figure*}

In this work we provide such a recipe and introduce the Perceiver-Actor-Critic (PAC) model which is outlined in~\Cref{fig:pac-teaser}.
With this, we show that \emph{a specific class of offline RL algorithms (offline actor-critic methods) can indeed scale to large models and datasets without incurring a large additional computational cost} and in addition we etablish, for the first time, that they \emph{follow similar scaling laws to those observed in the supervised learning regime}~\citep{henighan2020scaling,kaplan2020scaling}.
We further establish that this class of methods is ideally suited for slowly moving away from supervised BC towards RL during training, allowing us to run large and compute-intensive experiments without fear of instability and to adapt our method depending on the quality of the data.

We introduce a simple offline actor-critic algorithm that optimises a KL-regularized RL objective and can be seen as a simplified variant of MPO/DIME \citep{abdolmaleki2018maximum,abdolmaleki2021dime}.
We find that regularizing the policy towards the data distribution (via BC) is sufficient to stabilize offline RL for large models and also allows convenient interpolation between BC and RL.
We additionally introduce architectural advances which enable training with RL at scale.
E.g.\ incorporating the action into the Q-function via cross-attention (allowing fast estimation of Q-values for multiple actions) and incorporating a large number of inputs via Perceiver-style cross-attention to learned latent variables; enabling training with many inputs of different modalities (text, proprioception, vision) while enabling inference of a large \qty{1}{B} parameter model at \qty{20}{Hz} on a local machine.

PAC outperforms BC on a number of benchmarks in continuous control, including outperforming Gato \citep{reed2022generalist} on Control Suite \citep{tunyasuvunakool2020} tasks and recovers expert performance from heterogeneous data in a real robot benchmark. This establishes that RL should be considered a viable alternative to BC for large policies.

\section{Background and Related Work}

\paragraph{Supervised Generalist Agents}
Several recent works have trained large transformer-based \citep{vaswani2017attention} generalist agents via BC by building on previous works in which control tasks were transformed into sequence prediction problems \citep{chen2021decision,janner2021offline}. Gato \citep{reed2022generalist} for example, was trained on tasks ranging from Atari games to robotics manipulation.
Subsequently, large generalist robotics agents \citep{brohan2022rt,bousmalis2023robocat,zitkovich2023rt} have been trained on large datasets with multiple tasks, object sets and embodiments,
and have been shown to generalize to new tasks and domains after fine-tuning \citep{bousmalis2023robocat,padalkar2023rtx}.
Perceiver-based networks with cross-attention \citep{jaegle2021perceiver} have also been applied to robotics to minimize computational demands when handling voxel observations \citep{shridhar2023perceiver,ze2023gnfactor}.
Finally, \citet{team2023octo} used multi-headed attention to predict outputs in a similar way to the cross-attention in our system.

\paragraph{Offline RL}
Offline RL methods \citep{levine2020offline,lange2012batch} learn from fixed datasets without online exploration. Unlike supervised algorithms, they can learn from suboptimal trajectories and thus more data.
However, they are at risk of issues like overoptimism for unseen state-action pairs. This is often addressed by regularizing the policy to stay close to the data \citep{peng2019advantage,wang2020critic,fujimoto2019off,wu2019behavior}. Like prior work \citep{abdolmaleki2021dime, fujimoto2021minimalist}, we combine a BC regularization term with an off-policy RL method.
Other offline RL methods penalize the value function \citep{kumar2020conservative} or prevent value propagation \citep{kostrikov2021offline} for unseen state-action pairs. While most offline RL works use relatively small benchmarks, recent ones have tackled challenging multi-task problems \citep{kumar2022offline} and pre-trained robotics generalists that can be fine-tuned to new tasks \citep{kumar2022pre}.
However, to our knowledge, only the recent Q-Transformer \citep{chebotar2023q} provides an example of a transformer trained with offline RL on larger datasets, albeit with a relatively small model. Our actor-critic-based approach is more naturally suited for extending BC-based methods and less computationally demanding. This allows us to explore much larger models and perform a scaling law analysis.

\paragraph{Scaling Law Analysis}
Our scaling law analysis mirrors analyses of large language models for which several studies have shown smooth power-law relations between model size and performance \citep{kaplan2020scaling,hoffmann2022chinchilla,henighan2020scaling}. Some recent works have also investigated scaling behavior of neural networks for online RL \citep{neumann2022scaling,hilton2023scaling} albeit with relatively small ($<$40M parameter) models. % Max seems to be 10^7.5 parameters in the second reference.
\citet{lee2022multi} analyzed how performance scaled with the number of parameters of Decision Transformer \citep{chen2021decision} style networks and includes plots for a CQL \citep{kumar2020conservative} offline RL baseline for models up to 200M parameters finding no favourable scaling. In contrast, we find scaling to work for actor-critic methods and provide a thorough scaling law analysis. Concurrent work also shows promising scaling behavior of model-based RL methods \citep{hansen2023td} for models up to 300M parameters in a multitask setup.

\section{Scalable Offline Actor-Critic Learning}
\label{sec:method}

We scale up offline actor-critic methods to large models.
To achieve this,
we adapt methods from the offline RL literature and present our proposed algorithm in \Cref{sec:method-alg}.
We adapt Perceiver-IO~\citep{jaegle2022perceiver} architecture to the actor-critic setting and present our model in \Cref{sec:method-arch}.

\subsection{Background and Notation}
\label{sec:method-background}
We consider learning in a multi-task Markov decision process (MDP), where at each time step $t$ the agent selects an action $a_t\in\mathcal{A}$ for its current state $s_t\in\mathcal{S}$,
receives a reward $r_{t+1}=R(s_t,a_t,\task)\in\mathcal{R}$ specific to the task $\tau\in\mathcal{T}$  and transits to the next state $s_{t+1}\sim p(\cdot|s_t,a_t)$.
We use the term state and multimodal observations interchangably, although the true environment state is often not fully observable.

An RL algorithm seeks to find a policy $\pi(a_t|s_t,\task)$ that maximizes the per-task discounted cumulative return
$\EXP_{p_{\pi}}\left[\sum_{t=0}^\infty \gamma^t R(s_t, a_t,\task) \right]$ under the trajectory distribution $p_{\pi}$ induced by the policy $\pi$.
The Q-function, the V-function and the advantage function are defined as:
$Q^{\pi}(s_t,a_t,\task)=\EXP_{p_{\pi}, s_k=s_t, a_k=a_t}\left[\sum_{k=t}^{\infty} \gamma^{k-t} R(s_k,a_k,\task) \right]$,
$V^{\pi}(s_t,\task)=\EXP_{a_t\sim\pi(\cdot|s_t,\task)}\left[ Q^{\pi}(s_t,a_t,\task) \right]$,
$A^{\pi}(s_t,a_t,\task)=Q^{\pi}(s_t,a_t,\task)-V^{\pi}(s_t,\task)$.
We assume the availability of an offline dataset $\mathcal{D}=\lbrace (s_t,a_t,s_{t+1},\task) \rbrace$, generated by following a behavior policy $b(a_t|s_t,\task)$, and access to either the reward function $R$ or reward annotations.

We also make use of behaviour cloning (BC) terms for training which can be formalized as minimizing $\EXP_{(s_t, \task) \in \mathcal{D}} \KL[b, \pi| s_t,\task ] = -\EXP_{\mathcal{D}} \log \pi(a_t|s_t, \task) + K_{BC}$ between the behavior policy $b$ that generated the dataset and the learned policy $\pi$ ($K_{BC}$ is a constant offset).
%\begin{equation}
%\begin{aligned}
%    L^\text{BC}(\pi)
%    = \EXP_{(s_t, \task) \in %\mathcal{D}} \KL[b, \pi| s_t,\task ] 
%    = -\EXP_{\mathcal{D}} \log \pi(a_t|s_t, \task). 
%\end{aligned}
%\end{equation}

\begin{figure*}[ht!]
    \centering
    \includegraphics[width=0.99\textwidth]{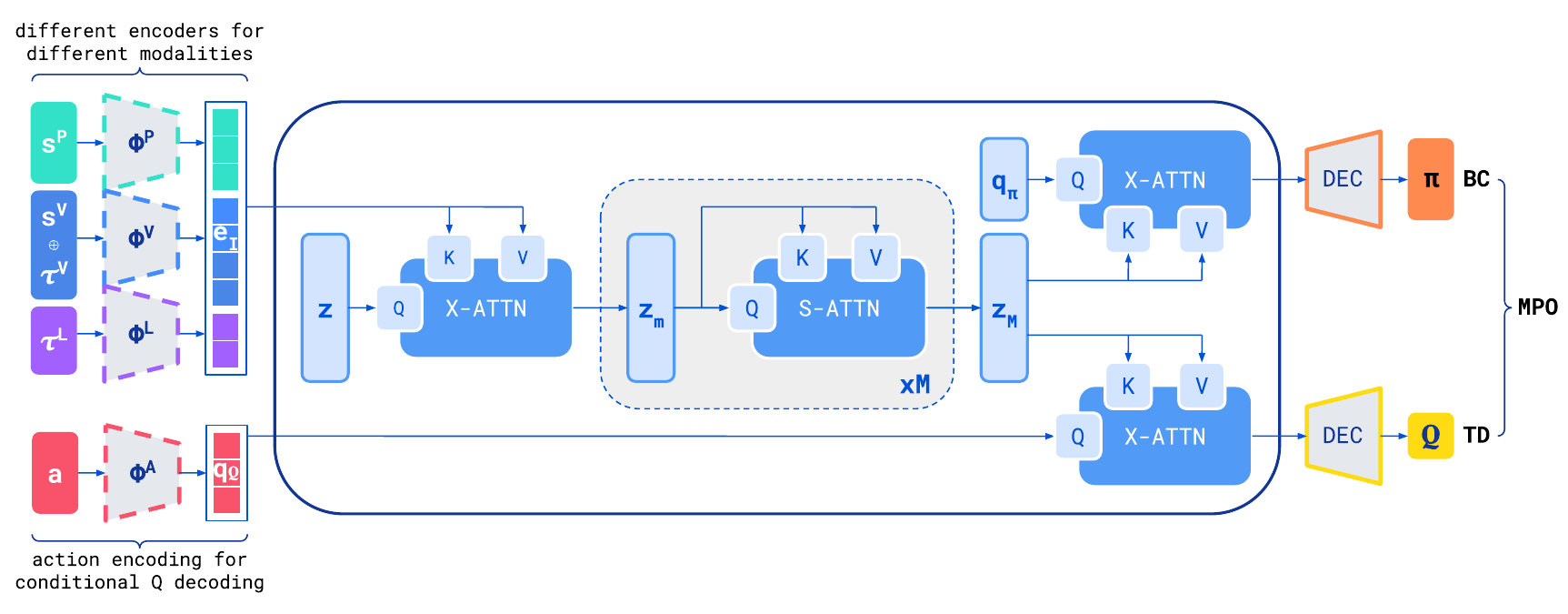}
    \caption{
    High-level PAC model architecture.
    Modality-specific encoders transform proprioceptive (P), visual (V), and language (L) inputs into embedding vectors $e_I$,
    which are cross-attended by learnable latent queries $z_0$.
    This is followed by a series of self-attention blocks to yield the latent encoding $z_M$,
    which is then queried via additional cross-attention modules to decode the desired outputs.
    The policy decoder employs a learnable query $q_{\pi}$ to cross-attend $z_M$ and outputs the logits of action distributions.
    The Q-value decoder employs a query $q_Q$ based on the encoded actions to cross-attend $z_M$ and outputs the action-specific logits of the distributional Q-function.
    }
    \label{fig:pac_arch}
\end{figure*}

\subsection{Offline KL-Regularized Actor-Critic}
\label{sec:method-alg}

We target a KL-regularized RL objective, where the goal is to find a policy $\piimp$ that improves over a reference policy $\pit$ via $\piimp = \arg \max_\pi J(\pi)$ where $J(\pi)$ is given as:
\begin{equation}
\begin{aligned}
    J(\pi)  
    &= \EXP_{(s_t, \task) \in \mathcal{D}}\left[\EXP_{a_t \sim \pi}\left[Q^\pi(s_t,a_t,\task)\right] - \eta \KL \left[ \pi,\pit|s_t,\task \right] \right] \\
    %&= \EXP_{(s_t, \task) \in \mathcal{D}}\left[\EXP_{a_t \sim \pi}\left[A^\pi(s_t,a_t,\task)\right] - \eta \KL \left[ \pi,\pit|s_t,\task \right] \right],
    %&= \EXP_{(s_t, \task) \in \mathcal{D}}\left[\EXP_{a_t \sim \pi(\cdot|s_t,\task)}\left[Q^\pi(s_t,a_t,\task)\right] - \eta \KL \left[ \pi,\pit|s_t,\task \right] \right] \\
    %&= \EXP_{(s_t, \task) \in \mathcal{D}}\left[\EXP_{a_t \sim \pi(\cdot|s_t,\task)}\left[A^\pi(s_t,a_t,\task)\right] - \eta \KL \left[ \pi,\pit|s_t,\task \right] \right],
\end{aligned}
\label{eq:imp_problem}
\end{equation}
where $\eta$ is a hyperparameter determining the strength of the regularization towards the reference policy $\pit$.
The solution to this maximization problem is given as (see~\Cref{sec:policy-optimality} for derivation): 
\begin{equation}
\begin{aligned}
\piimp(a_t | s_t, \task) &\propto \exp(\nicefrac{Q^{\piimp}(s_t,a_t,\task)}{\eta}) \pit(a_t | s_t,\task), \\
&\propto \exp(\nicefrac{A^{\piimp}(s_t,a_t,\task)}{\eta}) \pit(a_t | s_t,\task).
\end{aligned}
\label{eq:imp_policy}
\end{equation}
This observation allows us to transform the RL problem of finding an optimal policy into a weighted supervised learning problem (cf.~\citet{abdolmaleki2018maximum}). Assuming access to an estimate of $Q^{\piimp}$ or $A^{\piimp}$, we can fit a parametric policy $\pipac$ by minimizing its divergence $D_{KL}[\piimp,\pipac|s_t,\task]$ to $\piimp$ using a sample based estimate.
Turning the policy optimisation problem into an instance of supervised learning has the major benefit that it is easy to trade-off the policy optimisation objective with a behavior cloning term, since all loss terms are now (weighted) negative log likelihoods. 

Different choices for estimating $Q^{\piimp}$ or $A^{\piimp}$ as well as the reference policy $\pit$ lead to different algorithmic variants. We will concentrate on a Q-function based variant in the main paper but describe a state-value function (V-function) based variant in the appendix which has similar scaling benefits.

We train the policy $\pipac$ together with an estimate $Q_\theta \approx Q^{\pi_\theta} \approx Q^{\piimp}$ of the state-action value function.
To balance losses, we employ tools from the distributional reinforcement learning literature \citep{bellemare2017distributional} which transform the problem of learning $Q_\theta$ into minimizing the negative log likelihood of a discretized Q-function distribution $p_\theta(q | s_t, a_t, \task)$.
Using the distributional TD operator \citep{bellemare2017distributional} we can compute a sample-based target Q-distribution $\Gamma_{\theta'}(q | s_t, a_t, \task)$ (see \Cref{sec:pac-q-app}) where $\theta'$ are the parameters of a target network which is periodically updated to a time-lagged version of $\theta$. The same target parameters also give rise to a target policy $\pi_{\theta'}$ which we use as the reference policy in Equation \eqref{eq:imp_policy}, i.e. $\pit = \pi_{\theta'}$. Combining the policy loss, a BC loss, and the KL-based Q-value loss yields a total loss containing three KL terms:
\begin{equation}
\begin{aligned}
    L^Q(\theta) = \EXP_{\mathcal{D}}
    \Big[
    &(1 - \alpha) \KL[\piimp, \pipac | s_t,\task, \pit = \pi_{\theta'} ]  \\
    &+ \alpha \KL[b, \pipac | s_t,\task ]  \\
    &+ \beta \KL[\Gamma_{\theta'}(q| s_t,a_t,\task), p_\theta(q | s_t,a_t,\task)]
    \Big] \\
    = -\EXP_{\mathcal{D}}\Big[&(1 - \alpha) \EXP_{a' \sim \pi_{\theta'}}\left[w(a', s_t, \task) \log \pipac(a' | s_t,\task)\right] \\ 
    &+ \alpha  \log \pipac(a_t | s_t,\task) \\ 
    &+ \beta \EXP_{q \sim \Gamma_{\theta'}} \log p_\theta(q | s_t,a_t,\task) \Big] + K_H,
    \end{aligned}
    \label{eq:pac_q}
\end{equation}
where $w(a', s_t, \tau) = \frac{\exp(Q_\theta(s_t, a', \tau)/\eta)}{\EXP{a' \sim \pi_{\theta'}}[\exp(\nicefrac{Q_{\theta'}(s_t,a',\task)}{\eta})]}$ and $K_H$ is a constant entropy related offset independent of $\theta$.
The expectation over the data is estimated by sampling $(s_t, a_t, s_{t+1}, \task) \in \mathcal{D}$, the expectation over action samples from $\pi_{\theta'}$ is estimated based on $N = 10$ samples and the expectation $\EXP_{q \sim \Gamma_{\theta'}}$ can be evaluated analytically.
Finally $\alpha$ and $\beta$ are multipliers trading off different loss components (which are relatively easy to set due to all losses corresponding to weighted categorical log likelihoods). We refer to \Cref{sec:pac-q-app} for a step-by-step derivation.

Notably, aside from the KL towards the improved policy $\piimp$, Equation \eqref{eq:pac_q} also includes a KL towards the behaviour policy $b$.
This additional regularization is necessary to prevent $\pipac$ from converging to action samples that have high Q-values but are far away from those observed in the data (and are thus at the risk of being overestimated); a common issue in offline RL with Q-functions~\citep{levine2020offline}.
The additional BC term prevents this, following prior examples for using a BC loss as a simple regularisation technique in offline RL~\citep{abdolmaleki2021dime,fujimoto2021minimalist}.
We find that this is the only term needed to stabilize learning.
In addition, it gives us a natural way for moving away from learning via pure behavioral cloning ($\alpha = 1$) towards pure policy optimisation against the learned Q-function ($\alpha = 0$).
This also allows us to perform expensive training runs of large models with confidence since we can set $\alpha$ to a larger value such that the policy stays close to BC, guaranteeing stable training, and can reduce it later during fine-tuning.

\subsection{Scalable Architecture for Actor-Critic Learning}
\label{sec:method-arch}

With the proposed offline actor-critic algorithm, we now describe how $\pi_{\theta}$ and $Q_{\theta}$ are instantiated with scalable network architectures.  
In particular, we aim for an architecture that is flexible enough to incorporate different modalities of state observations and task descriptions as well as various action specifications, while also being computationally efficient for consuming high-dimensional inputs during learning and at inference time (to enable $\SI{20}{\hertz}$ control of real robots).
In this section, we describe how we adopt a Perceiver-IO architecture \citep{jaegle2021perceiver} to achieve the above.
The model is depicted in \Cref{fig:pac_arch}.

\paragraph{Observation Encoding}
\label{sec:arc-encoding}

Given multimodal inputs, in particular proprioceptive and visual observations $s_t=(s_t^P, s_t^V)$ along with visual and language task descriptions $\task=\task^V, \task^L)$,
our model first deploys one encoder ($\phi$) per modality to encode the inputs into embedding vectors:
$e_I = \phi^P(s_t^P) \oplus
\phi^V(s_t^V) \oplus
\phi^V(\tau^V) \oplus
\phi^L(\tau^L)
\in \mathbb{R}^{N \times D_I}$,
with $N$ and $D_I$ denoting the number and dimensionality of the embedding vectors.
Details of each modality encoder are provided in \Cref{sec:arc-encoding-app}.
For the proprioception encoder $\phi^P$ we propose a novel \textit{multi-scale} normalizer to account for arbitrary input scales and provide further details and ablations on this encoder choice in \Cref{sec:multi-scale-normalizer,sec:enc-ablations}.
We highlight that our model uses task descriptions of different modalities (text and vision) and we analyse this multimodal task conditioning in \Cref{sec:dropout}.

\paragraph{Transformer on Latent Space}
\label{sec:arc-latent}

At this point, the commonly adopted approach would be to feed the embedding sequence $e_I \in \mathbb{R}^{N \times D_I}$ directly into a transformer consisting of multiple stacked self-attention blocks.
However, for the domains we consider, the input sequence length amounts to thousands of tokens for a single time step.
As the computational complexity and memory usage of self-attention scales quadratically with the sequence length, this common treatment potentially hinders the learned controller from being applicable to real robotic systems that impose real-time constraints.
To address this, we adopt the methodology from the perceiver model \citep{jaegle2021perceiver}.
Specifically, a cross-attention block is placed at the front-end of the network in which the input sequence of embeddings $e_I$ are queried by $N_Z$ trainable latent vectors each of size $D_Z$: $z\in\mathbb{R}^{N_Z\times D_Z}$, which outputs latent embeddings $z_0$.
This is followed by $M$ self-attention operations on the latents which finally yield $z_M\in\mathbb{R}^{N_Z\times D_Z}$.
Since the number of latent vectors is typically much smaller than the input sequence length ($N_Z \ll N$) and the self-attention operation is shifted from the input embeddings to the latent vectors, this effectively reduces the computation and memory usage to $O({N_Z}^2)$.
We provide more details on the perceiver backbone in \Cref{sec:arc-latent-app}.

\paragraph{Policy and Value Decoding}
\label{sec:arc-decoding}
To implement an actor-critic algorithm, the model needs to output both a Q-value estimate and an action prediction.
While the action prediction $\hat{a}_t$ can be directly modeled as a function of the inputs $(s_t,\tau)$ which are encoded into $e_I$ and thus $z_M$, the value estimate $Q_{\theta}(s_t,a_t,\tau)$ also depends on the action $a_t$ which is not encoded in $z_M$.
To obtain the two types of outputs we cross-attend the latent embeddings $z_M$ with dedicated queries.
While the queries for the policy are learned vectors, the Q-value queries are computed by encoding the action $a_t\in\mathbb{R}^{N^A}$ via our multi-scale normalizer.
This has the advantage that the model is less prone to ignoring the action compared to when the action would be presented as an input (a common problem when learning Q-values).
It also allows efficient evaluation of the Q-function for multiple action samples via caching of the action-independent latent $z_M$.
We provide more details in \Cref{sec:arc-decoding-app} and ablate the importance of the cross-attention for Q-value prediction in \Cref{sec:xattn-ablation}.

\section{Experiments}
\label{sec:experiments}
We present three sets of experiments investigating different aspects of PAC.
\Cref{sec:scaling} analyzes whether PAC follows scaling laws similar to established supervised learning settings.
\Cref{sec:exp-pretraining} compares PAC's performance after large-scale training with the RL objective to different BC baselines across over 100 continuous control tasks.
Finally,~\Cref{sec:exp-rlft} studies how PAC can be finetuned by leveraging its Q-function to hone in on a real robot task and further improve its performance using self-generated data.

We use a large dataset throughout all experiments which combines tasks from three different sources:
Gato data~\citep{reed2022generalist} consist of records of an RL agent solving 32 simulation tasks in Control Suite~\citep{tunyasuvunakool2020}.
RoboCat data~\citep{bousmalis2023robocat} operates on the RGB Stacking benchmark~\citep{lee2021beyond} using RL in simulation to build pyramid and tower structures using a 7-DoF Panda robot.
It also contains an Insertion task featuring teleoperated simulation data of the same robot inserting differently sized gears onto pegs.
Lastly, CHEF~\citep{lampe2023mastering} data contains simulated and real-world records of a 5-DoF Sawyer robot stacking two objects in the RGB Stacking benchmark using an RL algorithm.
For all episodes in our dataset, a short language instruction describing the task is added to each frame, e.g. \texttt{humamoid.run} or \texttt{panda.sim.pyramid}, which serves as a unique goal instruction to differentiate between the different tasks. For all RoboCat tasks an additional goal image is provided as the goal instruciton. We again emphasize that our model can handle both language and visual goal descriptions (where present) and refer to \Cref{sec:dropout} for details about the goal conditioning.
In total, our data mix consists of 3.64M episodes across 102 simulated and 30 real continuous control tasks which equates to approximately 2.45T tokens for model training (cf.~\Cref{sec:exp-details-scaling,sec:tokenization}).

\begin{figure*}[t!]
    \begin{tabular}{c}
        \includegraphics[width=0.35\textwidth]{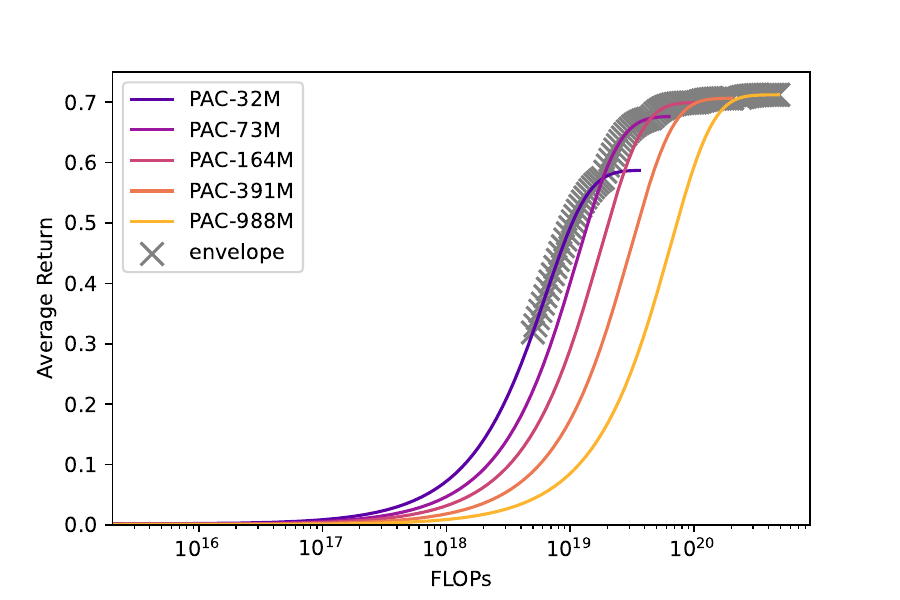}
        \includegraphics[width=0.31\textwidth]{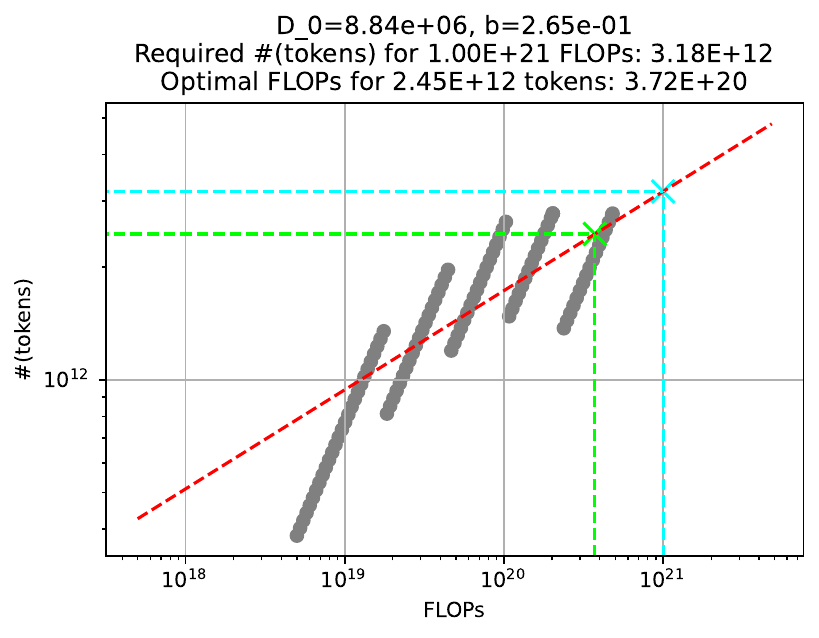}
        \includegraphics[width=0.31\textwidth]{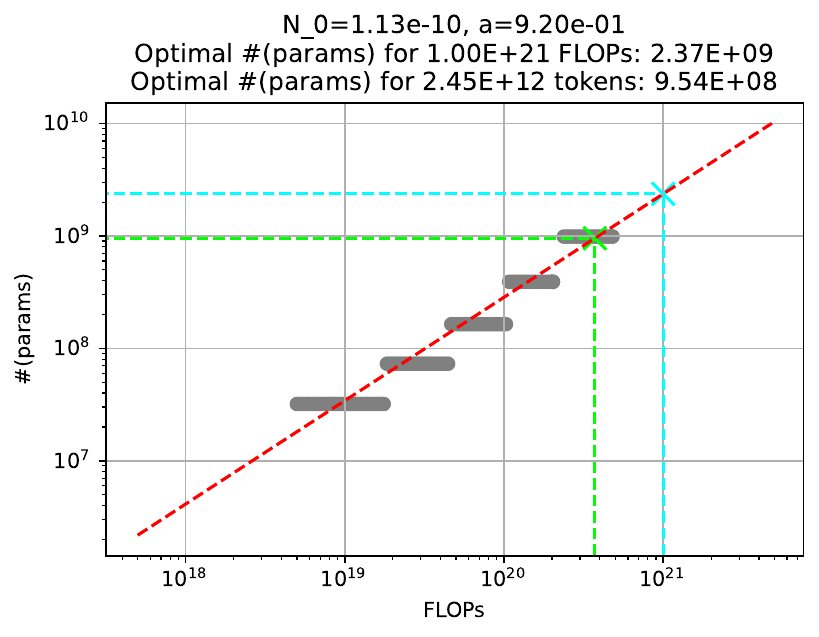} \\
    \end{tabular}
    \caption{
    Scaling laws based on the return profile envelope for \PACQ.
    We select 100 logarithmically spaced points between 5E+18 and 5E+20 FLOPs on the envelope of the return profiles (left) for the scaling law fits.
    For both the token and parameter scaling plots (middle, right), we indicate the scaling trend with a dashed red line.
    The green intersection represents the optimality point when training on a single epoch of our data while the teal intersection represents the optimal data and parameter trade-off for a FLOP budget of 1E+21.
    }
    \label{fig:qpi_reward_scaling}
\end{figure*}

\subsection{Scaling Analysis for Offline RL Objectives}
\label{sec:scaling}

A central part of our investigation is to understand the interaction between offline actor-critic algorithms and scalable neural network architectures that use (self-)attention.
When trained with supervised objectives, such as next-token prediction, architectures of this type usually follow \emph{scaling laws}~\citep{kaplan2020scaling}, i.e.\ for all performance-optimal models the number of tokens consumed and the number of model parameters used follow power-laws in the number of FLOPs spent.
However, it has so far been unclear whether these scaling laws also extend to RL.
To investigate this relationship, we adopt the methodology from~\citet{hoffmann2022chinchilla} (also known as `Chinchilla scaling laws') and apply it to PAC.
We define five different model scales (XXS, XS, S, M and L) ranging from 32M to 988M parameters to study the scaling behavior of PAC and report the full model architecture hyper-parameters in~\Cref{sec:model-scales}.

To conduct our analysis, we train PAC across the different scales with two different values of $\alpha$ for the BC/RL trade-off.
Setting $\alpha=1.0$ results in a BC objective for the policy and constitutes our baseline \BC\footnote{Using a Q-value loss term with $\beta>0$ never decreased the performance in our BC experiments; we keep it for comparability.} while \PACQ~performs offline RL with $\alpha=0.75$.
With a batch size of 512 trajectories of length five, one epoch of our data mix takes approximately 2.7M steps.
Therefore we train each model for 3M updates to stay in a single-epoch regime.

Following~\citet{kaplan2020scaling,hoffmann2022chinchilla}, the power laws between compute operations $C$, number of tokens $D$ and number of parameters $N$ for performance-optimal models of the family are:

\vspace{-0.3cm}
\begin{equation}
    N(C) = N_0 * C^a,~~~D(C) = D_0 * C^b .
    \label{eq:chinchilla-scaling-app1}
\end{equation}

Normally, the coefficients $a$ and $b$ are fitted using compute-optimal model checkpoints along the \emph{loss envelope} of the different training runs for different compute budgets.
However, we observe that the training loss is not a reliable indicator for model performance in our setting (cf.~\Cref{sec:loss-scaling}).
We therefore use an approximation of the average episode return as a means to select the best performing model for each compute budget from the respective model family.
To extrapolate from average returns we fit a logistic function to regress the training steps against average return across all tasks, normalized in $[0, 1]$ (cf.~\Cref{sec:return-profile-fitting}) to obtain a \emph{return profile} for each model.
We plot the return profiles for the \PACQ~family against FLOPs in the left column of~\Cref{fig:qpi_reward_scaling} and use them to select 100 points on the profiles' envelopes to fit the scaling laws of~\Cref{eq:chinchilla-scaling-app1}.
Scaling plots for all model families are presented in~\Cref{fig:reward_scaling_ext} in the Appendix.

\begin{figure}[h!]
    \begin{tabular}{c}
        \includegraphics[width=0.95\columnwidth]{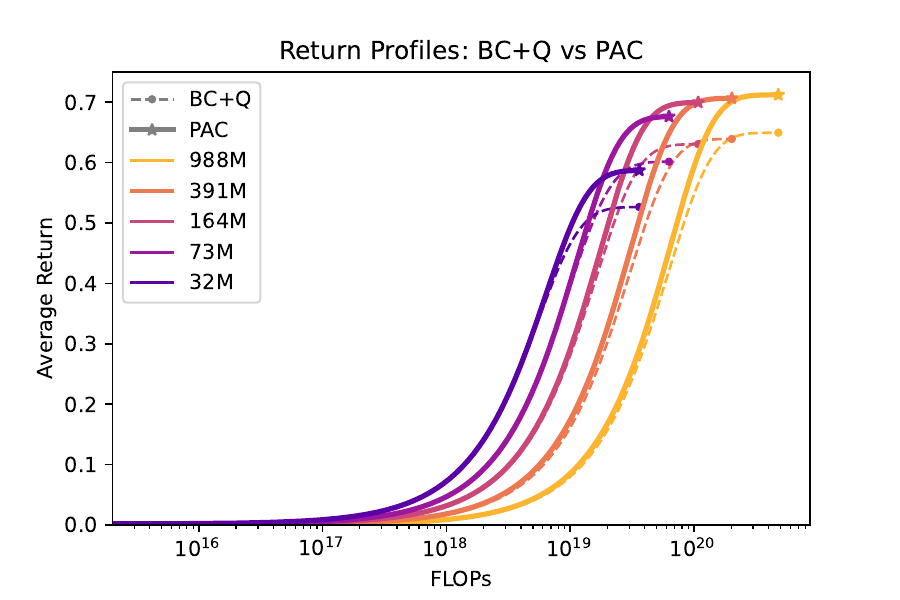} \\
        \includegraphics[width=0.95\columnwidth]{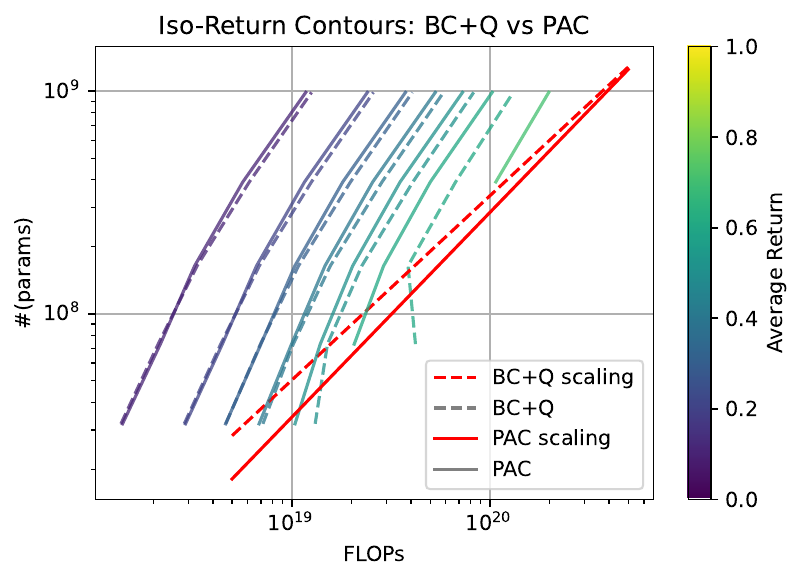} \\
    \end{tabular}
    \caption{
    Iso-Return comparison of \BC~vs \PACQ.
    The return profile (top) contrasts the expected average return between the BC baseline and the RL objective across all model scales.
    The Iso-Return contours (bottom) depict how the reward landscape over the parameter-FLOPs landscape shifts between using the BC objective (dashed contours) and the RL objectives (solid contours).
    }
    \label{fig:iso_reward_cmp}
\end{figure}

\begin{table*}[t]
\centering
\caption{
Policy success rates across $\#(\mathcal{T})$ tasks in each domain for 100 evaluations per task.
The average success rate in the training data is reported as $p_D$.
For Gato:Control, the percentage of achieved expert average reward and the standard-error-based 95\% CIs are reported.
For all other domains, the average success rates and their corresponding Wilson score intervals for $\alpha_W=0.05$ are reported.
Best results (within CI of the best mean) in each row are bold.
\scriptsize[$\dagger$ cited from \citet{reed2022generalist}; $\filledstar$ cited from \citet{bousmalis2023robocat}]
}
\label{tab:pretraining-results}
\begin{tabular}{lcc|ccc|cc}
\textbf{Domain} & $\#(\mathcal{T})$ & $p_D$ & \textbf{BC (Gato$\dagger$ / RC$\filledstar$)} & \textbf{FilteredBC} & \textbf{BC+Q} & \textbf{PAC} & \textbf{$\alpha$-PAC} \\
\hline
Gato:Control & 32 & N/A & 63.6$\dagger$ & 75.8~\tiny[62.5, 78.6] & 84.6~\tiny[79.6, 89.7] & 87.7~\tiny[83.8, 91.6] & \textbf{92.1~\tiny[88.4, 95.9]} \\
\hline\hline
RC:Tower & 7 & 75 & 61.0$\filledstar$~\tiny[57.3, 64.5] & 64.0~\tiny[60.4, 67.5] & \textbf{71.3~\tiny[67.8, 74.5]} & \textbf{69.3~\tiny[65.8, 72.6]} & \textbf{69.6~\tiny[65.9, 72.7]} \\
RC:Pyramid & 30 & 75 & \textbf{64.5$\filledstar$~\tiny[62.8, 66.2]} & \textbf{64.0~\tiny[62.3, 65.7]} & 62.4~\tiny[60.7, 64.1] & \textbf{63.5~\tiny[61.7, 65.1]} & \textbf{64.9~\tiny[63.1, 66.6]} \\
RC:Insertion & 3 & 97 & 71.3$\filledstar$~\tiny[66.0, 76.2] & 81.0~\tiny[75.8, 84.7] & 79.7~\tiny[74.8, 83.8] & 80.3~\tiny[75.5, 84.4] & \textbf{89.3~\tiny[85.0, 92.1]} \\
CHEF:sim & 1 & 28 & N/A & 17.0~\tiny[10.9, 25.5] & 11.0~\tiny[6.3, 18.6] & \textbf{55.0~\tiny[45.2, 64.4]} & \textbf{52.0~\tiny[42.3, 61.5]} \\
\end{tabular}
\end{table*}

The scaling laws are different for the BC and offline RL settings.
When we constrain the data budget to a single epoch, i.e.\ 2.45T tokens, the fits suggest to train a 1.33B parameter model in the \BC~case whereas in the case of \PACQ~a smaller model of only 954M parameters is suggested.
This is consistent with our observation that the L-size of \PACQ~with 988M parameters performs best which is close to the predicted optimality point while the \BC~model likely would benefit from being scaled up further.
Data-wise, \BC~and \PACQ~scale nearly the same ($b(\PACQ) \approx b(\BC) \approx 0.266$).
However, the RL objective seems to benefit more from additional parameters as the compute budget increases compared to BC ($a(\PACQ)=0.920 > a(\BC)=0.828$) suggesting that the capacity needed for the Q-function is larger (though as we will see the Q-function can learn from lower quality data).

Another way to compare the scaling behaviors between the BC and offline RL objectives is through the lens of the Iso-Return contours (analogous to the Iso-Loss landscape of~\citet{hoffmann2022chinchilla}) as presented in~\Cref{fig:iso_reward_cmp}.
The comparison between the isolines for different average return levels between the model families reveals an interesting pattern:
Compared to the BC baseline, the RL objective shifts the reward landscape to the top left of the parameters-FLOPs space.
This indicates that offline RL yields higher rewards for the same compute budget compared to BC.
Interestingly, the shift in the return isolines becomes more pronounced as the compute budget is increased which suggests that offline RL \emph{scales better} with increased compute than BC w.r.t.\ final task performance since the reward landscape ascends more steeply along the `efficient frontier' indicated by the parameter scaling law~\citep{hoffmann2022chinchilla}.

\subsection{Large-scale Offline Actor-Critic Learning}
\label{sec:exp-pretraining}
The scaling analysis above suggests that PAC's offline RL objective outperforms a BC objective as the compute is scaled up.
We now investigate whether this still holds when comparing against two strong BC baselines: Gato~\citep{reed2022generalist} and RoboCat~\citep{bousmalis2023robocat}.
The pre-training phase of such large models typically only uses a BC objective to ensure `safe' optimization and reduce the risk of divergence for these cost-intensive training runs.
However, if an offline RL objective could be used safely in the initial stage of training a large model, this would allow to train on sub-optimal data from the start and already learn a Q-function which can be leveraged for subsequent self-improvement.

For our comparison, we consider the following PAC-based models: \PACQ~($\alpha=0.75$) our main actor-critic model; \BC~($\alpha=1, \beta>0$) as a baseline which also learns a Q-function, but never leverages it for policy optimization (we found this to always be at least as good as pure BC in preliminary experiments); and \fBC~($\alpha=1, \beta=0$) which does not learn a Q-function and is only trained on successful episodes of our data mix to mimic a `pure' BC setting.
We also add \aPAC~as our best actor-critic model which uses a different value for the BC/RL trade-off $\alpha$ for each dataset to obtain the best performance and demonstrate that our method can be optimally tuned to deal with data of widely varying quality in the same training mixture.
More detailed ablations on the choice of $\alpha$ and $\beta$ are presented in \Cref{sec:hparam-sensitivity}.
For a fair comparison to the 1.2B parameter versions of Gato and RoboCat, we use PAC in its L-size with about 1B parameters and train for 3M updates.
All details of the pre-training data mix and optimizer hyper-parameters are reported in~\Cref{sec:exp-details-pretraining}.
Each PAC model is evaluated across all task families in simulation and the results are reported in~\Cref{tab:pretraining-results}.
Where available, we cite the baseline results for Gato and RoboCat (RC) directly from their respective papers.
In general, the Q-function-based PAC outperforms BC across tasks, confirming our hypothesis that offline RL is a viable alternative for training large models and we note that a V-function based variant also achieves similar results (see~\Cref{sec:pacv-pretraining}).

In more detail: On the Control Suite tasks \PACQ~outperforms all baseline tasks reaching $87.7\%$ of expert performance and \aPAC~even boosts it further to $92.1\%$.\footnote{For compatibility we also use the expert performance definition of \citet{reed2022generalist}.}
It is also worth noting that our BC baselines already outperform the Gato results, potentially due to PAC's improved architecture.
On the RoboCat tasks, \PACQ~performs commensurately with all BC baselines and outperforms prior work especially on the more difficult Tower task achieving $\approx 70\%$ success rate, but the difference is less pronounced since the respective datasets come from near expert policies ($>75\%$ success).
The biggest difference is observed on the insertion task where \fBC~and \BC~already improve $\approx 10\%$ over the RoboCat baseline and \aPAC~yields another significant improvement to $89.3\%$.
Finally, for the stacking task from CHEF which has the poorest data quality -- collected form a sub-optimal policy that only achieved $~28$\% success -- we can observe that \PACQ~learns policies with good success rates while all BC baseline are barely able to match the average performance of the data collecting policy.
This highlights that our method fulfills one of the main promises of offline RL: it can learn successful policies even from severely sub-optimal data.

\begin{table}[t]
\tabcolsep=0.12cm
\centering
\caption{
Success rates with Wilson score intervals for $\alpha_W=0.05$ for CHEF:real tasks (400 trials per task) for different objectives, as well as for an RL finetuning run with self-generated data (RLFT).
% Note, performance on other tasks did not degrade (not shown).
}
\begin{tabular}{lc|ccc}
\textbf{Domain} & $\#(\mathcal{T})$ & \textbf{BC+Q} & \textbf{\aPAC}  &  \textbf{\aPAC (RLFT)} \\
\hline
CHEF: real & 5 & 7.1 & 69.8 & \textbf{93.2} \\
~ & ~ & \tiny[6.1, 8.2] & \tiny[67.8, 71.8] & \textbf{\tiny[92.0, 94.2]}
\end{tabular}
\vspace{-0.25cm}
\label{tab:rlft-results-robot}
\end{table}

\subsection{RL Fine-tuning and Self-improvement}
\label{sec:exp-rlft}

We now demonstrate how PAC's built-in critic can be leveraged to transition into different finetuning scenarios and use this to `master' a target task (i.e.\ success rate $>90\%$).
For this we replicate the 5-DoF object stacking scenario of \cite{lee2021beyond} on a Rethink Sawyer arm in the real world.

Initially, we deploy different PAC models which have been pre-trained for 3M steps on the full data mix from~\Cref{sec:exp-pretraining}.
The best of these models (\aPAC) achieves a success rate of $69.8\%$ which far exceeds what is learnable from this data with BC (see \Cref{tab:rlft-results-robot}).
Additionally we verify that we can change the value of $\alpha$ during training, by first training with $\alpha = 1$ for 3M steps (cf.~\BC~in\Cref{tab:rlft-results-robot}) followed by 3M steps with $\alpha = 0$, which achieves $61.9\%~[60.0, 63.8]$, in line with the \aPAC result.
That demonstrates that we can safely transition from BC to RL at any point during the training process.

Next, we follow the iterative improvement protocol of~\citet{lampe2023mastering} and collect the evaluation trials in an additional dataset.
Afterwards, we add this data to the data mix (retaining all initial data used in previous sections) and train the model for another $\approx 250k$ steps.
We repeat this process multiple times, each time adding more data.
This cycle of feeding back self-generated data to the offline RL optimization provides a significant performance boost, increasing the success rate in each round, until eventually reaching a near-mastery level of $93.2\%$.
Average scores for each round, and the number of episodes collected for self-improvement, are summarized in \Cref{tab:robot-self-improvement}.
More detailed scores across the sub-tasks can be found in \Cref{sec:stacking-detail-results}.

Finally, we repeat this self-improvement experiment for all Control Suite tasks, adding $10,000$ episodes per task and performing RL finetuning for $500k$ steps starting from the checkpoint after three rounds of RLFT in \Cref{tab:robot-self-improvement}.
This results in an increase to $94.3\%~[91.3, 97.3]$, up from $92.1\%$ achieved by \aPAC.

The fine-tuning experiments highlight that PAC both outperforms BC on this challenging, low-quality data and can hill-climb its performance towards mastery using self-generated data -- a feat that is only possible with an RL style self-improvement loop.
Interestingly, even after mastering the CHEF:real domain, \aPAC's performance on the other domains does not decline as a side-effect (cf.~\Cref{tab:rlft-results-ext} in the Appendix).
It is also worth noting that the L-sized version of PAC runs at 20 Hz on a local Nvidia RTX 3090 GPU during this real-robot experiment.

\begin{table}[t]
\centering
\caption{
Success rates with Wilson score intervals for $\alpha_W=0.05$ for CHEF:real tasks (400 trials per task) across self-improvement iterations, as well as number of additional episodes collected for each iteration.
Rates are reported for the most challenging object flipping task (`set 2'), and the average across all test object sets ($\#(\mathcal{T})=5$).
}
\begin{tabular}{cr|cc}
\textbf{Iteration} & \textbf{Episodes} & \textbf{Flipping} & \textbf{CHEF:real} \\ \hline
\aPAC & 330k & 53.5~\tiny[48.6, 58.3] & 69.8~\tiny[67.8, 71.8] \\
RLFT \#1 & + 110k & 66.8~\tiny[62.0, 71.2] & 84.7~\tiny[83.1, 86.2] \\
RLFT \#2 & + 75k & 76.2~\tiny[71.8, 80.2] & 89.8~\tiny[88.4, 91.1] \\
RLFT \#3 & + 11k & 91.5~\tiny[88.4, 93.9] & 93.2~\tiny[92.0, 94.2] \\
\end{tabular}
\label{tab:robot-self-improvement}
\end{table}

\section{Discussion}

% conclusion
In this work, we demonstrated that offline actor-critic methods can scale to large models of up to 1B parameters and learn a wide variety of 132 control and robotics tasks.
On these tasks, our RL-trained models outperform strong BC baselines, especially in the presence of sub-optimal training data.
Our finetuning experiments also showed that RL can be effectively applied after pre-training without any model changes, which enabled the mastery of a real robot task improving from a $70\%$ to a $90\%$ success rate using RL and autonomously collected data.
The scaling analysis provides insights into the optimal model sizes and training durations for our datasets and indicates that the performance of offline RL scales better with compute than pure BC.
Finally, our system allows for a gradual and stable transition between BC and RL learning, and can process data of various modalities simultaneously, while remaining efficient enough to allow our biggest model to control a real robot at 20 Hz.

% limitations and future work
However, our work also has some limitations:
First, offline RL requires reward annotations, which can be costly.
Progress in the development of universal reward functions \citep{du2023vision} or unsupervised reward labeling \citep{chebotar2021actionable} could therefore greatly broaden the applicability of our method.
Second, given the wide variety of domains considered, we saw no strong indications of transfer across tasks.
However, we would expect generalization to improve with the use of datasets which have more overlap between tasks and domains as in \citet{zitkovich2023rt}.

Overall, we believe our work could pave the way for training large models via offline actor-critic methods on ever larger datasets of robot data.
Additionally, an exciting opportunity lies in further scaling offline actor-critic learning to models of multiple billion parameters, and combining our systems with pre-trained VLMs, or even exploring offline actor-critic RL as an alternative method for generative pre-training in language models.

\newpage

\section*{Broader Impact}
This work presents new methods for training generalist agents for control applications including robotic manipulation. 
The general impact on society from generalist robotics agents at this point is not well understood, and we encourage further work into their risks and benefits.
We emphasize that both our model and the actor-critic methods introduced for training at scale are for research use only and are not currently deployed in any production scenario to any uses, and we thus expect no direct impact resulting from this work.

In a broader sense, Perceiver-Actor-Critic shares the majority of safety concerns discussed in Gato \citep{reed2022generalist} and RoboCat \citep{bousmalis2023robocat}.
In particular, our self-improvement loop has the same safety concerns attached to the BC-style self improvement in \citet{bousmalis2023robocat}.
It is worth emphasising that our improvement step is carried out offline from human defined reward functions, and no learning happens while interacting with any real world system.
Additionally, in some sense the fact that we use rewards to 'shape' the behaviour of the learned policies makes work on safety via value alignment to human preferences \citep{russell2019human,christianorlhf} more directly applicable although much work still remains to be done on this front.

% \section*{Acknowledgements}
% stub

\bibliography{main}
\bibliographystyle{icml2024}

%%%%%%%%%%%%%%%%%%%%%%%%%%%%%%%%%%%%%%%%%%%%%%%%%%%%%%%%%%%%%%%%%%%%%%%%%%%%%%%
%%%%%%%%%%%%%%%%%%%%%%%%%%%%%%%%%%%%%%%%%%%%%%%%%%%%%%%%%%%%%%%%%%%%%%%%%%%%%%%
% APPENDIX
%%%%%%%%%%%%%%%%%%%%%%%%%%%%%%%%%%%%%%%%%%%%%%%%%%%%%%%%%%%%%%%%%%%%%%%%%%%%%%%
%%%%%%%%%%%%%%%%%%%%%%%%%%%%%%%%%%%%%%%%%%%%%%%%%%%%%%%%%%%%%%%%%%%%%%%%%%%%%%%
\clearpage
\appendix
%\onecolumn

\section*{Appendix}
This appendix presents further details and additional experiments of the proposed Perceiver-Actor-Critic (PAC) model and is arranged as follows:  
Methodological details are discussed in Sections \ref{sec:method-alg-app} and \ref{sec:method-arch-app},
with Section \ref{sec:method-alg-app} focusing on algorithmic details to complement Section \ref{sec:method-alg},
and Section \ref{sec:method-arch-app} discussing architecture details accompanying Section \ref{sec:method-arch}.
Then further details on the experimental setups are given in \Cref{sec:exp-details-app},
following which are algorithm-wise sensitivity analysis and architecture-wise ablation studies in Section \ref{sec:sensitivity-ablation-app}.
Finally, additional experiments are presented in \Cref{sec:additional-exp-app}.

\section{Method Details - Algorithm}
\label{sec:method-alg-app}

\subsection{Necessary Optimality Conditions for Policy}
\label{sec:policy-optimality}
We give a short derivation following previous work by \citet{abdolmaleki2018maximum}.
The solution to \Cref{eq:imp_problem} has to be optimal for every state $s$ and task $\tau$ and we will thus drop these dependencies for simplicity but add a Lagrangian because the policy is a probability distribution:
\begin{equation}
\begin{aligned}
    \piimp &= \arg \max_{\pi} J(\pi) \\
    &= \arg \max_{\pi} \EXP_\pi Q(a) - \eta \KL[\pi, \tilde \pi] + \lambda (1 - \EXP_\pi 1).
\end{aligned}
\end{equation}
We can now compute the partial derivative w.r.t. $\pi$ and set it to zero:
\begin{align}
    \partial J / \partial \pi &= Q(a) - \eta (\log(\pi(a)/\tilde \pi(a)) + 1) - \lambda = 0 \quad \forall a
\end{align}
After exponentiation this can be re-arranged into:
\begin{align}
    \exp(Q(a)/\eta) \exp(1+\lambda/\eta) = \pi(a)/\tilde \pi(a)
\end{align}
Since $\lambda$ is a constant and takes on the role of a normalizing constant we can write:
\begin{align}
    \pi(a) \sim \exp(Q(a)/\eta) \tilde \pi(a)
\end{align}
And \Cref{eq:imp_policy} can be retrieved by re-introducing the dependencies on states $s$ and task $\tau$.

\subsection{PAC+Q Details}
\label{sec:pac-q-app}
Our default implementation of PAC (called PAC+Q here for clarity) trains the policy $\pipac$ together with an estimate $Q_\theta \approx Q^{\pi_\theta} \approx Q^{\piimp}$ of the state-action value function. To enable us to balance losses easily we transform the problem of learning $Q_\theta$ into minimizing some negative log likelihood of a categorical distribution. This is possible by employing tools from the distributional reinforcement learning literature: Instead of parameterizing $Q_\theta$ directly, we learn a discretized representation: a categorical distribution $p_\theta(q | s_t, a_t, \task)$ over binned values $q \in \{q_\text{min}, q_\text{min} + \epsilon, q_\text{min} + 2 \epsilon, \ldots, q_\text{max}\}$ with bin size $\epsilon$, giving rise to the Q-value $Q_\theta(s_t, a_t, \task) = \EXP_{p_\theta} q$.

We use the definition of the distributional TD operator\footnote{Analogous to the standard temporal difference operator using the relation $Q^\pi(s_t, a_t) = r_t + \gamma E_\pi[Q^\pi(s_{t+1}, a_{t+1})].$} following \citet{bellemare2017distributional} to compute a target Q-distribution as:
\begin{equation}
\begin{aligned}
    &\Gamma_\theta(q \mid s_t,a_t,\task) =\\
    \EXP_{s_{t+1}}&\EXP_{\substack{a'\sim \\ \pi_{\theta}(\cdot \mid s_{t+1},\task)}}
    \EXP_{\substack{q'\sim \\ p_{\theta}(\cdot \mid s_{t+1},a',\task)}}
    \biggl[ 
        \mathbf{1}_{[q_t - \nicefrac{\epsilon}{2}, q_t + \nicefrac{\epsilon}{2}]}(r_t + \gamma q')
    \biggr],
\end{aligned}
\end{equation}
where the indicator function $\mathbf{1}$ is used to map the probability mass of the transformed Q-value to the nearest target bin. We can now compute the KL-divergence of our estimate against this target
\begin{align}
    \KL[\Gamma_{\theta'}(q| s_t,a_t,\task), p_\theta(q | s_t,a_t,\task)],
\end{align}
based on transitions $(s_t, a_t, s_{t+1}, \task) \in \mathcal{D}$ and where $\theta'$ refers to the parameters of a target network which we periodically update to a time-lagged version of model parameters $\theta$.

The above target network also gives rise to a target policy $\pi_{\theta'}$ and this is what we use as reference policy for the Q-value based maximisation of Equation \eqref{eq:imp_policy}, i.e. $\pit = \pi_{\theta'}$. Combining the policy loss, a BC loss, and the Q-Value loss into a total loss yields three KL terms:
\begin{equation}
\begin{aligned}
    L^Q(\theta) = &\EXP_{\mathcal{D}}
    \biggl[
    (1 - \alpha) \KL[\piimp, \pipac | s_t,\task, \pit = \pi_{\theta'} ] \\
    &+ \alpha \KL[b, \pipac | s_t,\task ] \\
    &+ \beta \KL[\Gamma_{\theta'}(q| s_t,a_t,\task), p_\theta(q | s_t,a_t,\task)]
    \biggr] \\
    = -\EXP_{\mathcal{D}}\biggl[&(1 - \alpha) \EXP_{a' \sim \pi_{\theta'}}\left[e^{\nicefrac{Q_{\theta'}(s_t,a',\task)}{\eta} - K_Q} \log \pipac(a' | s_t,\task)\right] \\
    &+ \alpha  \log \pipac(a_t | s_t,\task)  \\
    &+ \beta \EXP_{q \sim \Gamma_{\theta'}} \log p_\theta(q | s_t,a_t,\task) \biggr] + K_H,
    \end{aligned}
    \label{eq:pac_q_app}
\end{equation}
where $K_Q = \log \EXP_{a' \sim \pi_{\theta'}}[\exp(\nicefrac{Q_{\theta'}(s_t,a',\task)}{\eta})]$ is a normalizing constant, $K_H$ a constant entropy related offset independent of $\theta$, the expectation over the data is estimated by sampling $(s_t, a_t, s_{t+1}, \task) \in \mathcal{D}$, the expectation over action samples from $\pi_{\theta'}$ is estimated based on $N = 10$ samples and the expectation $\EXP_{q \sim \Gamma_{\theta'}}$ can be evaluated analytically.
Finally $\alpha$ and $\beta$ are multipliers trading off different loss components (which are relatively easy to set due to all losses corresponding to categorical log likelihoods).

\subsection{PAC+V Details}
\label{sec:pac-v-app}
Our alternative implementation of PAC uses a state-value function and is called PAC+V here for clarity.
It directly uses the data generating behavior policy as the reference, i.e. $\pit = b$, thus simplifying the losses.
Instead of estimating $Q^\pi$ we can thus rely on an empirical estimate of the advantage
$A^\pi(s_t,a_t,\task) \approx A_\theta(s_t,a_t,\task) = r_t + \gamma V_\theta(s_{t+1},\task) - V_\theta(s_t,\task)$ using samples $(s_t,a_t,s_{t+1},\task) \in \mathcal{D}$
from which we learn a V-function estimate $V_\theta \approx V^\pi$ only
As an aside: this also eliminates the need for a target network.

Analogously to the Q-function we use a categorical distribution $p_\theta(v|s_t,\task)$ over binned values $v \in \{v_\text{min}, v_\text{min} + \epsilon, v_\text{min} + 2 \epsilon, \ldots, v_\text{max}\}$ (note that this is now not conditioned on actions $a$) yielding the value estimate $V_\theta(s_t,\task) = \EXP_{p_\theta} v$. We can again compute target Q-values using a distributional TD operator:
\begin{equation}
\begin{aligned}
    &\Gamma_\theta(q| s_t,a_t,\task) = \\
    &\EXP_{s_{t+1}} \EXP_{v' \sim p_\theta(\cdot | s_{t+1},\task)}
    \left[ \mathbf{1}_{[q - \nicefrac{\epsilon}{2}, q + \nicefrac{\epsilon}{2}]}(r_t + \gamma v')
    \right].
\end{aligned}
\end{equation}
In order to retrieve target values we can take the expectation over our policy. However, given that we only have access to transitions from a behavior policy we require an importance weighting based correction:
\begin{align}
    \Gamma_\theta(v| s_t,\task) = \EXP_{a_t \sim b(\cdot|s_t,\tau)}
    \frac{\pipac(a_t | s_t,\task)}{b(a_t | s_t,\task)} \Gamma_\theta(q| s_t,a_t,\task).
\end{align}
% Since the value is the expectation However, this TD-operator estimates the value $V^b$ and not $V^\pi$ and thus we need to use importance sampling to correct for it, which can be achieved by minimizing the importance weighted divergence:
% \begin{equation}
%     \frac{\pipac(a_t | o^\task_t)}{b(a_t | o^\task_t)} \KL[\Gamma_{\theta'}^b(v| o^\tau_t, a_t, o^\tau_{t+1}), p_\theta(v | o^\task_t)].
% \end{equation} 

Using the second definition from Equation \eqref{eq:imp_policy} and choosing the reference to be the behavior policy $b$ as mentioned above we can then define the total loss for the V-function based actor-critic via the two KL terms:
\begin{equation}
    \begin{aligned}
    &L^V(\theta) = \EXP_{\mathcal{D}}
    \biggl[
    \KL[\piimp, \pi_\theta | s_t,\task, \pit = b] \\
    &\hspace{1.3cm}+ \beta \KL[\Gamma_{\theta'}(v| s_t,\task), p_\theta(v | s_t,\task)] \biggr] \\
    &= -\EXP_\mathcal{D}\biggl[ \exp\left(\nicefrac{(r_t + \gamma V_{\theta'}(s_{t+1},\task) - V_{\theta'}(s_t,\task))}{\eta} \right) \log \pipac(a_t | s_t,\task)\\
    &+ \beta \frac{\pipac(a_t | s_t,\task)}{b(a_t | s_t,\task)} \EXP_{v \sim \Gamma_{\theta'}(\cdot| s_t,a_t,\task)} \log p_\theta(v | s_t,\task) \biggr] + K_H,
    \end{aligned}
\end{equation}
where $K_H$ is again a constant entropy related offset and where we dropped the normalization constant for the exponentiated advantage as is common (see e.g. \citet{peng2019advantage}).
Even though we do not use target networks here, we keep the notation $\theta'$ to indicate that we do not take gradient w.r.t. the corresponding term.
Given that the behavior policy is used as reference for the policy improvement we do not need to include an additional BC term here -- this improvement step reverts to BC for $\eta \rightarrow \infty$ -- which is enough to ensure stable learning without overestimation issues. The only additional complication in this equation is that we need to estimate $b(a_t | s_t,\task)$ to compute importance weights. Two simple strategies are possible for this: we could either assume the behaviour policy executed actions from a fixed set, i.e. $b(a_t | s_t,\task) = \text{constant}$ in which case it can be dropped, or we can learn an estimate of $b$ via maximum likelihood (BC).

\section{Method Details - Architecture}
\label{sec:method-arch-app}

\subsection{Multi-scale Normalizer}
\label{sec:multi-scale-normalizer}
We propose \emph{multi-scale} normalizer that maps an input float of an arbitrary scale to an $N_G$-dimensional normalized representation: 
$\phi^{\text{multi-scale}}: \mathbb{R} \rightarrow [-1, 1]^{N_G}$. 
It does so by employing $N_G$ fixed gains across a logarithmic scale
(e.g. $\sigma = [10^{-4}, 10^{-3}, \ldots, 10^2, 10^3]$ for $N_G=8$)
to generate $\tanh$-bounded embeddings of the form:
\begin{equation}
\begin{aligned}
    \phi^{\text{multi-scale}}(x) = [ 
        \tanh(\sigma_1 x), \ldots, \tanh(\sigma_{N_G} x)
    ] .
    % ] \in \mathbb{R}^{N_G}.
\label{eq:multi-scale-app}
\end{aligned}
\end{equation}
For each input value the multi-scale normalizer thus generates a multi-dimensional representation vector using gains on a logarithmic scale.
Following this,
the attention mechanism,
for which the multi-scale normalizer is specifically designed to pair with,
can determine the most suitable gain levels to attend to.
% This scale can be chosen to cover a wide range, and in conjunction with attention mechanisms is well-suited for automatically determining the scale of an input observation.
This is in contrast to the commonly used $\mu$-Law scaling (e.g. in \citet{reed2022generalist}) for encoding continuous values which is at risk of either insufficient scaling or saturation if the input data is not within a known range.

\subsection{Observation Encoding}
\label{sec:arc-encoding-app}

As introduced in Section \ref{sec:method-background}, the state $s$ and the task description $\task$ can both be composed of different modalities: proprioceptive measurements (P),
visual observations (V)
and language descriptions (L), all of possibly domain-dependent variable sizes; and the actions (A) are treated as just another input modality.
Those input modalities are indicated by superscripts in notations.
Note that we assume, without loss of generality, that the task description $\task$ is episode-dependent and thus is denoted without a subscript indicating time step.
Also note that the task description for an offline episode can be sampled and relabeled in hindsight~\citep{andrychowicz2017her,riedmiller2018learning}.
So, we make the dependence on $\task$ explicit in notations.
In our experiments we allow for vision- and language-based task descriptions.

While easily extendable to other input data modalities, our current format accommodates proprioceptive and visual observations ($s^P$, $s^V$) as well as visual and language task descriptions ($\tau^V$, $\tau^L$):
\begin{itemize}
    \item Proprioception observations $s^P \in \mathbb{R}^{N^P}$ with ${N^P}$ denoting the dimensions of proprioceptive measurements.
    \item Visual observations $s^V \in \mathbb{R}^{N^V\times H \times W \times C}$ with number of image observations $N^V$, height $H$, width $W$, and number of channels $C$ for each image.
    \item Visual task descriptions $\tau^V \in \mathbb{R}^{N^V_{\tau}\times H \times W \times C}$ with $N^V_{\tau}$ number of images depicting the desired task (e.g., the last frame from a successful episode).
    \item Language task descriptions $\tau^L \in \mathbb{R}^{N^L_{\tau}}$ with number of text tokens $N^L_\tau$.
\end{itemize}
All input modalities across all domains are padded in order to match this format and we keep track of the valid entries with a mask.
The complete input at time step $t$ is therefore the concatenation of
$(s_t=(s^P_t, s^V_t), \tau=(\tau^V, \tau^L))$.
This could also be easily extended to incorporate more than a single timestep in case of partially observable tasks.

The concatenated input is then mapped via one encoder per modality into multiple embedding vectors of dimension $D_I$.
% That is, we apply a mapping $\phi(s_t,\tau)$ that maps the aggregated input into
% a sequence of embedding vectors of size
% $\mathbb{R}^{(N^P + N^V + N^V_{\tau} + N^L_{\tau}) \times D_I}$,
The details of each modality encoder are given below.
The output of this mapping will be fed into the perceiver encoder.

The proprioception encoder $\phi^{P}$ is an instantiation of the multi-scale normalizer (see \Cref{sec:multi-scale-normalizer}) followed by a linear layer $L^P$ with $D_I$ output dimensions to map the multi-scale representation to the desired shape:
$\phi^{P} = L^{P}\cdot\phi^{\text{multi-scale}}_{N_G}: \mathbb{R} \rightarrow \mathbb{R}^{N_G \times D_I}$.

The image inputs are encoded via a ResNet~\citep{he2016resnet}.
We omit the last projection layer to obtain $N_E$ spatial dimensions for our image embedding $\phi^V: \mathbb{R}^{H\times W\times C} \rightarrow \mathbb{R}^{N_E\times D_I}$,
where $N_E$ depends on the input image size and the down-sampling specifications of the ResNet.
We note that just like the proprioception embedding, and in contrast to other approaches, we do not discretize or tokenize image inputs but use continuous mappings instead.

We assume language inputs to be tokenized by the Sentence-Piece tokenizer~\citep{kudo2018sentencepiece} during data loading.
These are then directly embedded using a learnable look-up table $\phi^L: [1 .. N_T] \rightarrow \mathbb{R}^{D_I}$, with $N_T$ the total number of different language tokens. 

Applying each encoder to the corresponding input modality thus generates an encoded input,
$e_I = \phi^P(s_t^P) \oplus
\phi^V(s_t^V) \oplus
\phi^V(\tau^V) \oplus
\phi^L(\tau^L)
\in \mathbb{R}^{N \times D_I}$,
with $N = N^P N_G + N^V N_E + N^V_{\tau} N_E + N^L_{\tau}$.

\subsection{Transformer on Latent Space}
\label{sec:arc-latent-app}

Most state-of-the-art large-scale models for control take the encoded input
$e_I \in \mathbb{R}^{N \times D_I}$
and directly feed it into a transformer consisting of multiple stacked self-attention blocks.
For the domains in our experiments and the data modalities that we consider, this input sequence would be of length
$N=2,634$
for a single time step (cf.~\Cref{sec:tokenization}).
As the computational complexity and memory usage of the self-attention mechanism scales quadratically with the input sequence length,
this commonly adopted treatment potentially hinders the learned generalist controller from being applicable to real robotic systems that impose real-time constraints,
and more generally restricts efficient learning and inference,
let alone providing feasible solutions to extend the temporal context window beyond a few time steps or include more high-dimensional inputs.
To address this challenge,
we adopt the methodology from the perceiver model \citep{jaegle2021perceiver}
to shift the operation of stacked self-attention from the input sequence onto a few trainable latent vectors,
which in turn reduces the computational complexity from a quadratic dependence on the input sequence length to a linear one.

For ease of discussion,
we first give an overview of the attention operator
($f_{\text{ATTN}}$)
and the attention block (ATTN),
which is the main building block of the perceiver model.
The attention operator takes as inputs three matrices:
the queries
$Q\in\mathbb{R}^{N_Q\times D_Q}$,
the keys
$K\in\mathbb{R}^{N_K\times D_Q}$
and the values 
$V\in\mathbb{R}^{N_K\times D_V}$,
where the queries and the keys match in their vector dimensionality ($D_Q$)
while the keys and values contain the same number of vectors ($N_K$).
The attention operator is then defined as
$f_{\text{ATTN}}(Q,K,V)=\text{softmax}\left(\nicefrac{QK^T}{\sqrt{D_Q}}\right)V\in\mathbb{R}^{N_Q\times D_V}$,
for which the majority of the computational complexity and memory usage scale with $O\left(N_Q\cdot N_K\right)$
(the dependency of the scaling factor on the vector dimensionality is left out to keep the discussion concise).
Combining the attention operator $f_{\text{ATTN}}$ with linear projection layers,
normalization layers,
post processing MLPs and skip connections
yields the attention block
$\text{ATTN}(x_Q,x_K,x_V)$
(we refer to \citep{vaswani2017attention,jaegle2021perceiver} for more details).
Note that this most general form admits possibly three different sources of inputs to be linearly projected to act as
$Q$,
$K$ and
$V$ respectively:
$x_Q\in\mathbb{R}^{N_Q\times N_1}$,
$x_K\in\mathbb{R}^{N_K\times N_2}$,
$x_V\in\mathbb{R}^{N_K\times N_3}$
(there is no constraint of $N_1$ matching $N_2$ here as the three input sources will each go through a linear projection before being fed into $f_{\text{ATTN}}$).

Getting back at the encoded input $e_I\in\mathbb{R}^{N\times D_I}$,
a standard transformer would instantiate the attention block in the form of self-attention
(using the same data source for the queries as for the keys-and-values):
$\text{S-ATTN}(e_I)=\text{ATTN}(x_Q=e_I,x_K=e_I,x_V=e_I)$.
With $e_I$ containing $N$ vectors of dimentionality $D_I$,
the computation required by the self-attention blocks will scale quadratically with the input sequence length: $O\left(N^2\right)$,
which imposes challenges for meeting real-time control constraints or enabling efficient learning and inference.

We address this issue with a perceiver-style backbone.
Instead of applying self-attention directly on the input sequence,
$N_Z$ learnable vectors are created with each being of size $D_Z$.
We refer to those trainable vectors as the latents
$z\in\mathbb{R}^{N_Z\times D_Z}$.
The number of latent vectors is typically much smaller than the length of the input sequence:
$N_Z \ll N$
(e.g., $N_Z=32$ in our experiments).
With this,
we instantiate the attention block in another form,
namely a cross-attention between the latents and the normal inputs:
$\text{X-ATTN}(z,e_I)=\text{ATTN}(x_Q=z,x_K=e_I,x_V=e_I)$,
where the latents are used as the data source for the queries and the input sequence for the keys-and-values. 
This cross-attention block thus scales linearly with the input sequence length: $O(N_Z\cdot N)$,
and is positioned at the front-end of our PAC model (cf.~\Cref{fig:pac_arch}),
incorporating information from the inputs into the latent vectors.
Following this,
a standard transformer stack composed of $M$ self-attention blocks is applied on the latent space:
$z_{m+1}=\text{S-ATTN}(z_{m})$,
which finally outputs an encoding
$z_M\in\mathbb{R}^{N_Z\times D_Z}$.
We note that the quadratic computational scale of those self-attention blocks is decoupled from the input sequence length to be
$O({N_Z}^2)$.
This effectively shifts conducting the self-attention operation from the inputs to the latents,
in our case reducing the computation and memory usage by a factor of $(\nicefrac{2,634}{32})^2 \approx 6,775$.

\subsection{Policy and Value Decoding}
\label{sec:arc-decoding-app}
To obtain the two types of outputs,
we create dedicated query vectors for each output which are then used to query the latent encodings $z_M$;
querying information from $z_M$ is again implemented via cross-attention 
following the decoding architecture of Perceiver-IO \citep{jaegle2022perceiver})
and deriving keys and values from the latents $z_M$.

Specifically,
to acquire an action prediction of shape
$N^A\times N_B$,
with $N^A$ denoting the cardinality of the action space $|\mathcal{A}|$
and $N_B$ the number of bins each element gets discretized into,
we create $N^A$ learnable policy queries
$q_{\pi}\in\mathbb{R}^{N^A\times N_O}$
to cross-attend to $z_M$,
$\text{X-ATTN}(q_{\pi},z_M)$,
and get a $N^A\times N_O$ shaped output,
which are then linearly projected to $N^A\times N_B$ shaped policy logits.

Whereas for the Q-value estimate,
since the information about the action are not contained in the encoded latents $z_M$ but required for getting the estimate,
the Q-value queries $q_Q$ should not be simply created as randomly initialized trainable vectors as for $q_{\pi}$.
Instead,
they are computed by encoding the action $a_t\in\mathbb{R}^{N^A}$ via an action encoder $\phi^A$ composed of an multi-scale normalizer 
(
$\mathbb{R}^{N^A}\rightarrow[-1,1]^{N^A\times N_G}$,
cf.~\Cref{eq:multi-scale-app}
)
followed by two linear projections
(
% $L^{A_1}:\mathbb{R}^{N_G} \rightarrow \mathbb{R}^{N_G\times D_O}$,
$L^{A_1}:{[-1,1]}^{N^A\times N_G} \rightarrow \mathbb{R}^{(N^A\times N_G)\times D_O}$,
$L^{A_2}:\mathbb{R}^{(N^A\times N_G)\times D_O} \rightarrow \mathbb{R}^{1\times D_O}$
):
$\phi^A=L^{A_2}\cdot L^{A_1}\cdot \phi^{\text{multi-scale}}_{N_G}:\mathbb{R}^{N^A}\rightarrow\mathbb{R}^{1\times D_O}$.
The generated value query
$q_Q\in\mathbb{R}^{1\times D_O}$
contains only one vector of size $D_O$,
since for each action the Q-value estimate only needs to output one quantity.
Note that in constrast to how the observation encoders
($\phi^P$, $\phi^V$, $\phi^L$)
are introduced in Section \ref{sec:arc-encoding-app}
as mappings from each individual element to the corresponding encoding,
here the action encoder $\phi^A$ is denoted as the mapping from the full action dimension $N^A$ to its corresponding encoding.
$q_Q$ is then used to query the latents via cross-attention,
$\text{X-ATTN}(q_Q,z_M)$,
the output of which
($1\times N_O$)
is then mapped to
$1\times N_Q$
to generate the $N_Q$ logits.
Incorporating the action information by encoding it into a query at the decoding stage
instead of concatenating it along with the other observations at the input
has the advantage that this way the action is less likely to be ignored by the model
(a common problem encountered when learning Q-values).
It also allows efficient evaluation of the Q-function for multiple action samples:
the latent representation $z_M$ is not dependent on the action and therefore needs to be computed only once to be queried by multiple action samples.

\section{Experimental Details}
\label{sec:exp-details-app}

\subsection{Architecture Hyperparameters for Scaling}
\label{sec:model-scales}

We vary three parameters of the architecture as detailed in~\Cref{tab:pac-scaling-hypers}:
The size of the latent vectors $D_Z$, the number of self-attention blocks \texttt{M} and the widening factor \texttt{W} of the attention blocks which define the ratio between the residual MLPs' hidden size to input size.
% All other, fixed hyperparameters of the perceiver IO backbone are reported in~\Cref{sec:perceiver-hparams}.
With all the other fixed hyperparameters reported in~\Cref{sec:perceiver-hparams},
the resulting model sizes range from 32M parameters (comparable to RT-1~\citep{brohan2022rt} and Q-Transformer~\citep{chebotar2023q}) to 988M parameters (close to the largest versions of Gato~\citep{reed2022generalist} and RoboCat~\citep{bousmalis2023robocat}.

\begin{table}[h]
\centering
\caption{Hyperparameters of PAC's different model scales.}
\label{tab:pac-scaling-hypers}
\begin{tabular}{cc|ccc}
\textbf{Scale} & \#(params) & $D_Z$ & \texttt{M} & \texttt{W} \\ \hline
\textbf{XXS} & 32M & 768 & 4 & 1 \\
\textbf{XS} & 73M & 1024 & 8 & 1 \\
\textbf{S} & 164M & 1280 & 10 & 2 \\
\textbf{M} & 391M & 1536 & 12 & 4 \\
\textbf{L} & 988M & 2048 & 18 & 4 \\
\end{tabular}
\end{table}

\subsection{Fixed Architecture Hyperparameters}
\label{sec:perceiver-hparams}

\Cref{tab:pac-perceiver-hypers} provides an overview of all employed model parameters which are kept fixed across model scales.
$N^P$, $N^V$, $N^V_{\tau}$, $N^L_{\tau}$, $N^A$ all refer to input dimensions and are chosen to accommodate all tasks the data is originating from. $N_E$ and $N_T$ also relate to input dimensions, but depend on the pre-processing that is applied to the data (e.g. the SentencePiece tokenizer for text input, or a ResNet and for image input).
$N_G$ is the number of scales (gains) for our proposed multi-scale normalizer.
Finally, $N_B$, refers to the number of discrete value bins used for the discretization of the action and $N_Z$ refers to the number of latent vectors.

\begin{table}[h]
\footnotesize
\centering
\caption{Constant hyperparameters of PAC across all model scales.}
\label{tab:pac-perceiver-hypers}
\begin{tabular}{l|c}
\textbf{Hyperparameter} & \textbf{Value} \\ \hline
$N^P$: proprioception dimensions & 223 \\
$N^V$: image observations & 5 \\
$N^V_{\tau}$: goal images & 3 \\
$N^L_{\tau}$: text tokens for task descriptions & 50 \\
$N^A$: action dimensions & 38 \\
$N_G$: multi-scale normalizer scales & 8 \\
% $S_{min}, S_{max}$: input scale bounds & $10^{-4}, 10^3$ \\
$N_E$: visual tokens per image & 100 \\
$N_T$: language token vocabulary size & 32,000 \\
$N_Z$: number of Perceiver latents & 32 \\
$N_B$: action bins & 101 \\
$D_I$: token embedding size for each modality & 256 \\
\end{tabular}
\end{table}

% NOTE(ogroth): Table here for layouting purposes.
\begin{table*}[ht!]
  \begin{center}
    \caption{
    The data mixture used for the scaling experiments.
    For each domain we report the modalities (P = proprioception, V = vision, L = language, A = actions), the number of tasks, the number of episodes recorded, the effective number of trajectories (each consisting of five timesteps), the number of tokens contributed, the percentage of successful episodes and the weight $\lambda$ of the domain during the data sampling process.
    \scriptsize[* = success rate only available for five stacking tasks; ** = success rate average only computed across 82 tasks with available success rates]
    }
    \label{tab:datasets-summary}
    \begin{tabular}{l|ccccccc}
      \textbf{Domain} & \textbf{Mod} & $\#(\mathcal{T})$ & \textbf{\#(Ep)} & \textbf{\#(Traj)} & \textbf{\#(Tok)} & \textbf{success} & $\lambda$ \\
      \hline
      Gato: Control & P, L, A & 32 & 468k & 187M & 430B & 47\% & 8 \\
      CHEF: sim & P, V, L, A & 30 & 755k & 60M & 430B & 28\%* & 2 \\
      CHEF: real & P, V, L, A & 30 & 2M & 169M & 1.12T & 12\%* & 2 \\
      RC: Tower & P, V, L, A & 7 & 121k & 19M & 122B & 72\% & 1 \\
      RC: Pyramid & P, V, L, A & 30 & 194k & 31M & 195B & 52\% & 1 \\
      RC: Insertion & P, V, L, A & 3 & 100k & 40M & 154B & 97\% & 2 \\
      \hline
      \multicolumn{2}{r}{$\sum$ =} & 132 & 3.638M & 506M & 2.45T & avg = 42\%** \\
    \end{tabular}
  \end{center}
\end{table*}

\subsection{Data Modalities and Tokenization}
\label{sec:tokenization}

\paragraph{Images}
At each timestep, PAC processes up $N^V + N^V_\tau$ input images.
For domains which provide fewer observations or goal images, these inputs are zero-padded and masked out for the subsequent processing.
Across all our experiments, we use an image resolution of 80~$\times$~80 pixels.
Each image is encoded using a simple ResNet~\citep{he2016resnet} with three blocks using (32, 64, 64) channels, $3 \times 3$ pooling kernels and $2 \times 2$ striding.
After the convolutions, each image has been downsampled to 10~$\times$~10 `image tokens' which are projected into an embedding space of $D_I$ dimensions.
Under this embedding scheme, each image is counted as 100 tokens in our experiments.
Our default implementation uses eight image observations at each timestep resulting in the processing of 800 image tokens.

\paragraph{Proprioception and Actions}
Our \emph{multi-scale normalizer} (cf.~\Cref{sec:multi-scale-normalizer}) represents each floating point number of the proprioceptive reading as $N_G$ tokens.
Each of these tokens is then projected into an embedding space of $D_I$ dimensions.
Our default implementation uses 223 proprioception observations at each timestep resulting in the processing of 1,784 proprioception tokens.
In domains with fewer proprioception observations, the remaining inputs are zero-padded and masked out for the subsequent processing.

\paragraph{Language}
In order to differentiate between different tasks in the same domain, e.g. to tell the model whether to execute a \texttt{run}, \texttt{stand} or \texttt{walk} policy in Control Suite's \texttt{humanoid} domain, we feed a simplified task instruction as a language task description ($\task^{L}$) at each timestep.
For each dataset, the task instruction is constructed from the dataset name (cf.~\Cref{tab:data-mixture}) and looks like \texttt{humanoid.walk} or \texttt{sim.insert.large\_gear}.
We use the SentencePiece tokenizer~\citep{kudo2018sentencepiece} to tokenize the task instruction where each token represents an integer index into a vocabulary of $N_T$ tokens.
Each language token is subsequently projected into an embedding space of $D_I$ dimensions.
In our default implementation, we use at most 50 language tokens for the task instruction and zero-pad and mask language tokens in cases of shorter instructions.

In summary, our default implementation processes a total amount of 2,634 input tokens per timestep broken down into: 800 image tokens, 1,784 proprioception tokens and 50 text tokens.

\subsection{Data Mixtures}
\label{sec:exp-details-scaling}
\begin{figure}[h!]
    \includegraphics[width=0.99\linewidth]{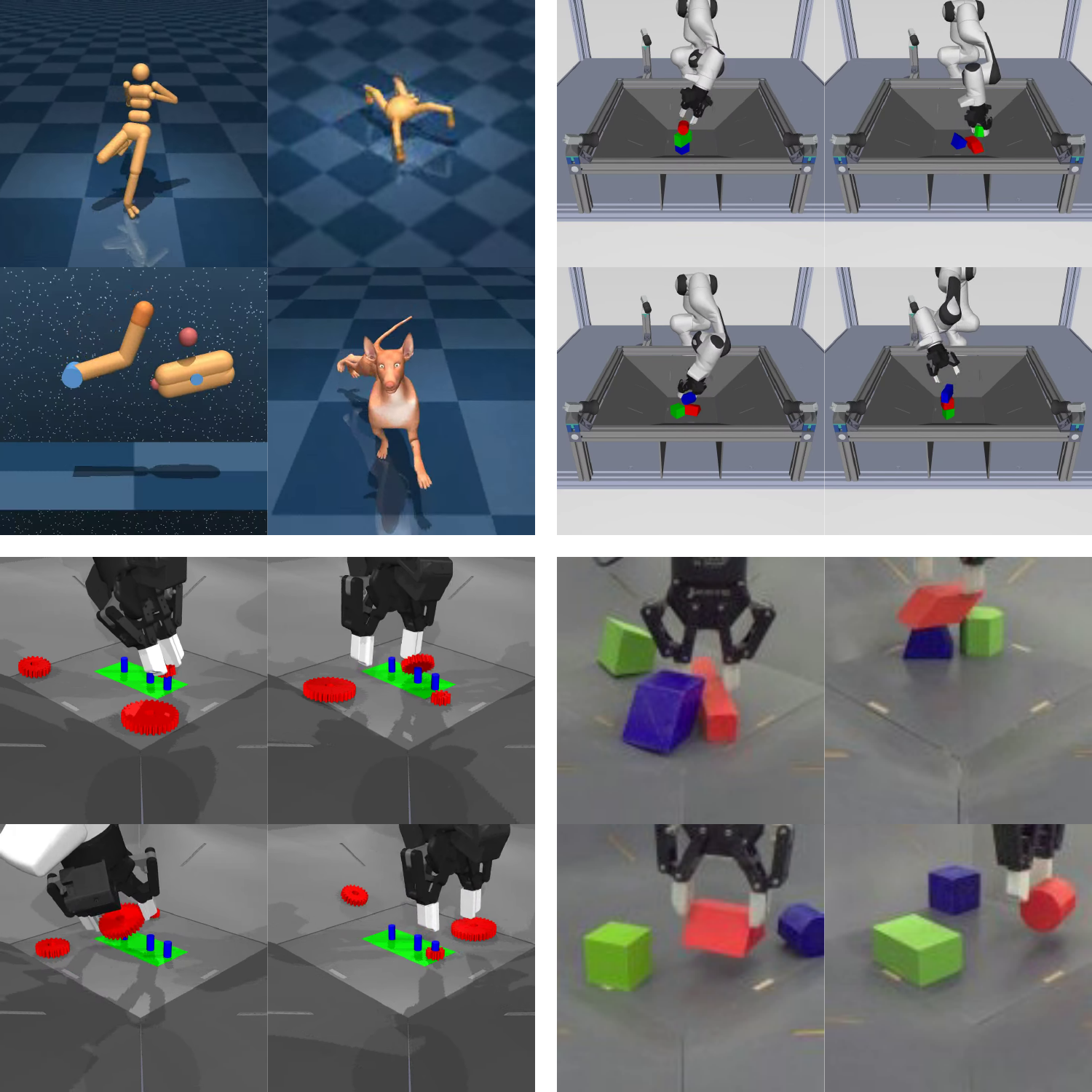}
    \caption{
    A selection of the domains and tasks in our data mix.
    \textbf{Top left:} Control Suite features 32 different continuous control tasks across 15 different embodiments with a great variance in proprioception and action spaces.
    \textbf{Top right:} Stacking RGB objects into different configurations (pyramides and towers) with a simulated Panda arm.
    \textbf{Bottom left:} Inserting gears onto pegs in simulation.
    \textbf{Bottom right:} Performing the RGB stacking task on a real Sawyer robot.
    }
    \label{fig:pac-domains}
\end{figure}

During our scaling experiments (cf.~\Cref{sec:scaling}), we use a data mixture consisting of 51 datasets and depict a selection of them in \Cref{fig:pac-domains}.
We partition the datasets into six groups as outlined in~\Cref{tab:datasets-summary}.
When sampling trajectories during training, we first sample uniformly across all groups with a weight of $\lambda$ and then sample uniformly within the group to draw a trajectory sample from a concrete dataset.
The effective sampling ratio for each dataset is shown as $P_{eff}$ in \Cref{tab:data-mixture}.
To maintain the sampling ratios over the entire training process, we simply loop all underlying datasets deterministically such that none of them is ever exhausted.
% estimated token per update: 9.26E+05 effective; 7.52E+06 padded

During our large-scale pre-training experiments (cf.~\Cref{sec:exp-pretraining}), we augment the data mixture already used in the scaling experiments (see the previous section), but add the full amount of RoboCat data for the Tower and Pyramid domains from~\citet{bousmalis2023robocat} to enable a fair comparison with prior work.
The adjusted overall data mixture is shown in \Cref{tab:datasets-summary-pt} and the changes to the individual datasets are reported in \Cref{tab:data-mixture-pt}.

\subsection{Optimization Hyperparameters}
\label{sec:exp-details-pretraining}

For all large-scale experiments (cf.~\Cref{sec:scaling,sec:exp-pretraining}) we use optimizer hyperparameters as reported in~\Cref{tab:pac-scaling-hypers-optimizer}.
Importantly, we use the AdamW optimizer~\citep{loshchilov2017adamw} with a learning rate schedule which starts at \texttt{lr\_init}, ramps up linearly for \texttt{lr\_warmup\_steps} to \texttt{lr\_peak} and is then cosine-annealed to \texttt{lr\_end} over \texttt{lr\_decay\_steps} which amounts to approximately one epoch in our data mix.
In line with the protocol of~\citet{hoffmann2022chinchilla}, we decay the learning rate by one order of magnitude over one epoch.
We set \texttt{lr\_decay\_steps} according to the respective epoch lengths for each of our experiments, i.e. to 2.7e6 in the scaling experiments and to 4.7e6 in the pre-training experiments.

For all \PACQ~models, we keep the TD loss scale $\beta$ constant at $38$  while varying the BC vs RL trade-off $\alpha$ between $1.0$ for our \BC~and \fBC~baselines and $0.75$ for the \PACQ~model series.
\aPAC sets $\beta=19$ and $\alpha$ of $0.75$ for the Control Suite and RoboCat data (i.e. leaning more towards BC given the high average success rates in the data) and $\beta=1,900, \alpha=0.0$ for the CHEF datasets (i.e. relying fully on the RL term in the presence of highly sub-optimal data).
All \PACV~models use $\alpha=0.0$, $\beta=38$ and a temperature $\tau$ of 1e-4.

\section{Sensitivity and Ablation Experiments}
\label{sec:sensitivity-ablation-app}

\subsection{Sensitivity to Hyperparameters}
\label{sec:hparam-sensitivity}
Here we report additional experiments that demonstrate the influence of some of the hyperparameters and that informed the settings that we used in other experiments.

\subsubsection{BC Loss Scale $\alpha$}
In order to perform offline reinforcement learning (RL), we simply interpolate between the behavioral cloning (BC) loss and the RL loss via a parameter $\alpha$.
We found that, depending on the nature of the data, different parameters work best for different datasets.
Empirically, we observed that the more `expert data' there is (i.e. a higher base success rate of the tasks in the dataset), the higher the contribution of the BC loss should be.
Conversely, less expert data can benefit more from RL.
As seen in \Cref{tab:bc_loss_scale}, for the control suite, averaged over 32 tasks, setting $\alpha$ at around 0.8 works best. This is because most of the data has been collected by expert agents and therefore small divergence from BC distribution can gain improvement over the BC.
This value is different for, for example, CHEF data, where $\alpha=0.0$ (i.e pure RL) works best mainly because the data expert level is low and this dataset contains more low quality data.

\begin{table}[h!]
\centering
\caption{
    Control Suite performance of an XS-sized model using different BC loss scales $\alpha$.
    The percentage of achieved expert average reward across 100 trials per task and the standard-error-based 95\% CIs are reported.
}
\label{tab:bc_loss_scale}
\tabcolsep=0.10cm
\begin{tabular}{lc|cccc}
  \textbf{Domain} & $\#(\mathcal{T})$ & \textbf{$\alpha$=0.0} & \textbf{$\alpha$=0.4} & \textbf{$\alpha$=0.8} & \textbf{$\alpha$=1.0} \\
\hline
Gato: Control & 32 & 36.8 & 74.3 & 85.1 & 82.4 \\
~ & ~ & \tiny[33.9, 39.6] & \tiny[70.2, 78.3] & \tiny[80.4, 89.8] & \tiny[76.5, 88.2] \\
\end{tabular}
\end{table}
\label{sec:exp-sensitivity-bc}

\subsubsection{Policy Loss Scale $\beta$}
For efficient learning we share the network parameters between the policy and critic networks.
This, however, requires some adjustments to balance these two losses for optimal performance.
Therefore, we introduce a hyperparameter, $\beta$, that trades off the policy loss and the critic loss.
To illustrate the sensitivity of this parameter, we use the CHEF tasks and perform pure reinforcement learning by setting BC loss scale $\alpha=0.0$ while sweeping over different values of $\beta$.
As can be seen in \Cref{tab:pi_loss_scale}, for this task, a lower contribution of the policy loss leads to better results.
In general, we have found that setting the $\beta$ value to 0.005 achieves good performance across various tasks.

\begin{table}[h!]
\centering
\caption{
    Simulated RGB stacking performance of an XS-sized model using different policy loss scales $\beta$.
    The average success rates across 100 trials per task and their corresponding Wilson score intervals for $\alpha_W=0.05$ are reported.
}
\label{tab:pi_loss_scale}
\tabcolsep=0.12cm
\begin{tabular}{lc|cccc}
 \textbf{Domain} & $\#(\mathcal{T})$ & \textbf{$\beta$=1.0} & \textbf{$\beta$=0.1} & \textbf{$\beta$=0.01} & \textbf{$\beta$=0.005} \\
\hline
CHEF: sim & 1 & 22.0 & 29.0 & 76.0 & 83.0 \\
~ & ~ & \tiny[15.0, 31.7] & \tiny[21.0, 38.5] & \tiny[66.8, 83.3] & \tiny[74.5, 89.1] \\
\end{tabular}
\end{table}

\subsection{Architecture Ablations}
\label{sect:architecture_ablations}
To understand what architectural decisions contribute to the performance we conduct a set of ablation studies.
In particular we look into the multi-scale normalizer for encoding continuous inputs,
the output action-cross-attention to generate Q-values,
and the task conditioning.
For compute efficiency reasons we perform these experiments on a subset of the data and with smaller model sizes.

\subsubsection{Input Encoding}
\label{sec:enc-ablations}
Our multi-scale normalizer is used for embedding both the proprioception observations and actions.
We compare the multi-scale normalizer against a vanilla linear encoder and a token-based encoder similar to the one used by \citet{reed2022generalist}.
The linear encoder directly embeds the inputs to a single embedding using a linear layer followed by a normalization layer.
The token-based encoder uses a lookup-table-based encoder analogous to the language encoder.
Across all tokenization variants, the embedded proprioception tokens are separately cross-attended while the embedded action tokens are projected to a single embedding vector.

\Cref{tab:ablations} reports separate results on the control suite tasks (using our XS-sized model) and RoboCat pyramid and tower tasks (using our S-sized model), where for better comparability with RoboCat \citep{bousmalis2023robocat} we set the number of action bins to $1024$.
For Control Suite the linear embedder shows the lowest performance, likely caused by not being able to deal with the range of proprioception measurements.
Based on the $\mu$-law discretization, encoder is able to address the high range of measurements.
However for the RoboCat tasks the $\mu$-law encoder performs the worst, possibly due to discarding neighborship relations in the data. It also comes with a larger memory footprint due to the lookup table embedding.
The multi-scale normalizer addresses all these issues and performs best.

%%% BEGIN: TABLE PAGE 1 %%%
\begin{table*}
\footnotesize
\centering
\caption{
    Data mixture of our scaling experiments.
    For each dataset, we report the number of trajectories and tokens contributed (\#(Traj) and \#(Tok)), the number of proprioception, vision and action dimensions in the raw data ($N^P$, $N^V$, $N^A$) and the probability $P_{eff}$ with which we sample from each dataset during training.
}
\label{tab:data-mixture}
\begin{tabular}{l|rrrrrr}
\textbf{Dataset Group} & \#(Traj) & \#(Tok) & $N^P$ & $N^V$ & $N^A$ & $P_{eff}$ \\ \hline
\texttt{Control Suite} \\
\texttt{~~acrobot.swingup} & 10,843,200 & 5.75E+09 & 6 & 0 & 1 & $1/64$ \\
\texttt{~~ball\_in\_cup.catch} & 5,292,800 & 3.44E+09 & 8 & 0 & 2 & $1/64$ \\
\texttt{~~cartpole.balance} & 10,473,600 & 5.13E+09 & 5 & 0 & 1 & $1/64$ \\
\texttt{~~cartpole.swingup} & 10,704,800 & 5.25E+09 & 5 & 0 & 1 & $1/64$ \\
\texttt{~~cartpole.three\_poles} & 10,012,800 & 7.31E+09 & 11 & 0 & 1 & $1/64$ \\
\texttt{~~cartpole.two\_poles} & 4,875,200 & 2.97E+09 & 8 & 0 & 1 & $1/64$ \\
\texttt{~~cheetah.run} & 7,672,800 & 8.12E+09 & 17 & 0 & 6 & $1/64$ \\
\texttt{~~dog.run} & 4,238,800 & 4.43E+10 & 223 & 0 & 38 & $1/64$ \\
\texttt{~~dog.stand} & 4,162,400 & 4.45E+10 & 223 & 0 & 38 & $1/64$ \\
\texttt{~~dog.trot} & 4,163,200 & 4.45E+10 & 223 & 0 & 38 & $1/64$ \\
\texttt{~~dog.walk} & 5,861,600 & 6.27E+10 & 223 & 0 & 38 & $1/64$ \\
\texttt{~~finger.spin} & 4,257,200 & 2.94E+09 & 9 & 0 & 2 & $1/64$ \\
\texttt{~~finger.turn\_easy} & 9,893,200 & 8.01E+09 & 12 & 0 & 2 & $1/64$ \\
\texttt{~~finger.turn\_hard} & 6,794,000 & 5.50E+09 & 12 & 0 & 2 & $1/64$ \\
\texttt{~~fish.swim} & 6,172,800 & 7.96E+09 & 21 & 0 & 5 & $1/64$ \\
\texttt{~~fish.upright} & 6,682,400 & 8.62E+09 & 21 & 0 & 5 & $1/64$ \\
\texttt{~~hopper.hop} & 7,054,000 & 7.12E+09 & 15 & 0 & 4 & $1/64$ \\
\texttt{~~hopper.stand} & 3,620,400 & 3.66E+09 & 15 & 0 & 4 & $1/64$ \\
\texttt{~~humanoid.run} & 4,632,000 & 1.75E+10 & 67 & 0 & 21 & $1/64$ \\
\texttt{~~humanoid.stand} & 5,181,200 & 1.95E+10 & 67 & 0 & 21 & $1/64$ \\
\texttt{~~humanoid.walk} & 7,566,800 & 2.85E+10 & 67 & 0 & 21 & $1/64$ \\
\texttt{~~pendulum.swingup} & 7,992,400 & 3.28E+09 & 3 & 0 & 1 & $1/64$ \\
\texttt{~~point\_mass.easy} & 3,980,400 & 1.95E+09 & 4 & 0 & 2 & $1/64$ \\
\texttt{~~quadruped.escape} & 3,102,000 & 1.48E+10 & 101 & 0 & 12 & $1/64$ \\
\texttt{~~quadruped.run} & 1,730,400 & 6.66E+09 & 78 & 0 & 12 & $1/64$ \\
\texttt{~~quadruped.walk} & 4,457,600 & 1.72E+10 & 78 & 0 & 12 & $1/64$ \\
\texttt{~~reacher.easy} & 5,501,200 & 3.14E+09 & 6 & 0 & 2 & $1/64$ \\
\texttt{~~swimmer.swimmer15} & 5,483,600 & 1.78E+10 & 61 & 0 & 14 & $1/64$ \\
\texttt{~~swimmer.swimmer6} & 98,000 & 1.42E+08 & 25 & 0 & 5 & $1/64$ \\
\texttt{~~walker.run} & 1,988,000 & 2.88E+09 & 24 & 0 & 6 & $1/64$ \\
\texttt{~~walker.stand} & 7,401,200 & 1.07E+10 & 24 & 0 & 6 & $1/64$ \\
\texttt{~~walker.walk} & 5,334,400 & 7.73E+09 & 24 & 0 & 6 & $1/64$ \\
\hline
\texttt{CHEF} \\
\texttt{~~sim.lift} & 10,072,080 & 7.16E+10 & 129 & 19,200 & 5 & $1/48$ \\
\texttt{~~sim.open} & 10,072,080 & 7.16E+10 & 129 & 19,200 & 5 & $1/48$ \\
\texttt{~~sim.place} & 10,072,080 & 7.16E+10 & 129 & 19,200 & 5 & $1/48$ \\
\texttt{~~sim.reach\_grasp} & 10,072,080 & 7.16E+10 & 129 & 19,200 & 5 & $1/48$ \\
\texttt{~~sim.stack} & 10,072,080 & 7.16E+10 & 129 & 19,200 & 5 & $1/48$ \\
\texttt{~~sim.stack\_leave} & 10,072,080 & 7.16E+10 & 129 & 19,200 & 5 & $1/48$ \\
\texttt{~~real.lift} & 28,224,012 & 1.87E+11 & 129 & 12,800 & 5 & $1/48$ \\
\texttt{~~real.open} & 28,224,012 & 1.87E+11 & 129 & 12,800 & 5 & $1/48$ \\
\texttt{~~real.place} & 28,224,012 & 1.87E+11 & 129 & 12,800 & 5 & $1/48$ \\
\texttt{~~real.reach\_grasp} & 28,224,012 & 1.87E+11 & 129 & 12,800 & 5 & $1/48$ \\
\texttt{~~real.stack} & 28,224,012 & 1.87E+11 & 129 & 12,800 & 5 & $1/48$ \\
\texttt{~~real.stack\_leave} & 28,224,012 & 1.87E+11 & 129 & 12,800 & 5 & $1/48$ \\
\hline
\texttt{RoboCat} \\
\texttt{~~panda.sim.triple\_stack.success} & 13,992,960 & 8.80E+10 & 44 & 51,200 & 7 & $1/32$ \\
\texttt{~~panda.sim.triple\_stack.failure} & 5,439,066 & 3.42E+10 & 44 & 51,200 & 7 & $1/32$ \\
\texttt{~~panda.sim.pyramid.success} & 15,998,400 & 1.01E+11 & 44 & 51,200 & 7 & $1/32$ \\
\texttt{~~panda.sim.pyramid.failure} & 15,024,082 & 9.45E+10 & 44 & 51,200 & 7 & $1/32$ \\
\texttt{~~sim.insert\_large\_gear} & 13,372,152 & 5.18E+10 & 21 & 32,000 & 7 & $1/24$ \\
\texttt{~~sim.insert\_medium\_gear} & 14,182,310 & 5.49E+10 & 21 & 32,000 & 7 & $1/24$ \\
\texttt{~~sim.insert\_small\_gear} & 12,260,934 & 4.74E+10 & 21 & 32,000 & 7 & $1/24$ \\
\end{tabular}
\end{table*}

\begin{table*}[h!]
  \begin{center}
    \caption{
    The data mixture used for the pre-training experiments.
    For each domain we report the modalities (P = proprioception, V = vision, L = language, A = actions), the number of tasks, the number of episodes recorded, the effective number of trajectories (each consisting of five timesteps), the number of tokens contributed, the percentage of successful episodes and the weight $\lambda$ of the domain during the data sampling process.
    \scriptsize[* = success rate only available for five stacking tasks; ** = success rate average only computed across 82 tasks with available success rates]
    }
    \label{tab:datasets-summary-pt}
    \begin{tabular}{l|ccccccc}
      \textbf{Domain} & \textbf{Mod} & $\#(\mathcal{T})$ & \textbf{\#(Ep)} & \textbf{\#(Traj)} & \textbf{\#(Tok)} & \textbf{success} & $\lambda$ \\
      \hline
      Gato: Control & P, L, A & 32 & 468k & 187M & 430B & 47\% & 4 \\
      CHEF: sim & V, P, L, A & 30 & 755k & 60M & 430B & 28\%* & 1 \\
      CHEF: real & V, P, L, A & 30 & 2M & 169M & 1.12T & 12\%* & 1 \\
      RC: Tower & V, P, L, A & 7 & 100k & 16M & 100B & 75\% & 2 \\
      RC: Pyramid & V, P, L, A & 30 & 601k & 96M & 604B & 75\% & 5 \\
      RC: Insertion & V, P, L, A & 3 & 100k & 40M & 154B & 97\% & 1 \\
      \hline
      \multicolumn{2}{r}{$\sum$ =} & 132 & 4.024M & 567M & 2.84T & avg = 58\%** \\
    \end{tabular}
  \end{center}
\end{table*}

%%% END: TABLE PAGE 1 %%%

%%% BEGIN: TABLE PAGE 2 %%%

\begin{table*}[h!]
\footnotesize
\centering
\caption{
    Data mixture of our scaling experiments.
    For each dataset, we report the number of trajectories and tokens contributed (\#(Traj) and \#(Tok)), the number of proprioception, vision and action dimensions in the raw data ($N^P$, $N^V$, $N^A$) and the probability $P_{eff}$ with which we sample from each dataset during training.
    Group entries abbreviated by $\dots$ are identical with the ones reported in \Cref{tab:data-mixture}.
}
\label{tab:data-mixture-pt}
\begin{tabular}{l|rrrrrr}
\textbf{Dataset Group} & \#(Traj) & \#(Tok) & $N^P$ & $N^V$ & $N^A$ & $P_{eff}$ \\ \hline
\texttt{Control Suite} \\
\texttt{~~}$\dots$ & ~ & ~ & ~ & ~ & ~ & $4/14$ \\
\texttt{CHEF} \\
\texttt{~~}$\dots$ & ~ & ~ & ~ & ~ & ~ & $2/14$ \\
\texttt{RoboCat} \\
\texttt{~~panda.sim.triple\_stack.success.eval\_set\_2} & 1,244,000 & 7.82E+9 & 44 & 51,200 & 7 & $1/42$ \\
\texttt{~~panda.sim.triple\_stack.failure.eval\_set\_2} & 681,818 & 4.29E+9 & 44 & 51,200 & 7 & $1/42$ \\
\texttt{~~panda.sim.triple\_stack.success.eval\_set\_4} & 7,363,360 & 4.63E+10 & 44 & 51,200 & 7 & $1/42$ \\
\texttt{~~panda.sim.triple\_stack.failure.eval\_set\_4} & 1,843,906 & 1.16E+10 & 44 & 51,200 & 7 & $1/42$ \\
\texttt{~~panda.sim.triple\_stack.success.eval\_set\_5} & 3,439,040 & 2.16E+10 & 44 & 51,200 & 7 & $1/42$ \\
\texttt{~~panda.sim.triple\_stack.failure.eval\_set\_5} & 1,366,208 & 8.59E+10 & 44 & 51,200 & 7 & $1/42$ \\
\texttt{~~panda.sim.pyramid.success.eval\_set\_1} & 15,024,000 & 9.45E+10 & 44 & 51,200 & 7 & $5/140$ \\
\texttt{~~panda.sim.pyramid.failure.eval\_set\_1} & 4,367,962 & 2.75E+10 & 44 & 51,200 & 7 & $5/140$ \\
\texttt{~~panda.sim.pyramid.success.eval\_set\_2} & 13,409,760 & 8.43E+10 & 44 & 51,200 & 7 & $5/140$ \\
\texttt{~~panda.sim.pyramid.failure.eval\_set\_2} & 5,763,134 & 3.63E+10 & 44 & 51,200 & 7 & $5/140$ \\
\texttt{~~panda.sim.pyramid.success.eval\_set\_3} & 13,249,600 & 8.33E+10 & 44 & 51,200 & 7 & $5/140$ \\
\texttt{~~panda.sim.pyramid.failure.eval\_set\_3} & 5,749,294 & 3.62E+10 & 44 & 51,200 & 7 & $5/140$ \\
\texttt{~~panda.sim.pyramid.success.eval\_set\_4} & 15,385,920 & 9.68E+10 & 44 & 51,200 & 7 & $5/140$ \\
\texttt{~~panda.sim.pyramid.failure.eval\_set\_4} & 3,767,426 & 2.47E+10 & 44 & 51,200 & 7 & $5/140$ \\
\texttt{~~panda.sim.pyramid.success.eval\_set\_5} & 15,318,240 & 9.64E+10 & 44 & 51,200 & 7 & $5/140$ \\
\texttt{~~panda.sim.pyramid.failure.eval\_set\_5} & 4,053,880 & 2.55E+10 & 44 & 51,200 & 7 & $5/140$ \\
\texttt{~~sim.insert\_large\_gear} & 13,372,152 & 5.18E+10 & 21 & 32,000 & 7 & $1/42$ \\
\texttt{~~sim.insert\_medium\_gear} & 14,182,310 & 5.49E+10 & 21 & 32,000 & 7 & $1/42$ \\
\texttt{~~sim.insert\_small\_gear} & 12,260,934 & 4.74E+10 & 21 & 32,000 & 7 & $1/42$ \\
\end{tabular}
\end{table*}

\begin{table*}[h!]
\centering
\caption{Optimizer hyperparameters for all model scales across all experiments.}
\label{tab:pac-scaling-hypers-optimizer}
\begin{tabular}{l|ccccc}
\textbf{Hyperparameter} & \textbf{XXS (32M)} & \textbf{XS (73M)} & \textbf{S (164M)} & \textbf{M (391M)} & \textbf{L (988M)} \\ \hline
\texttt{lr\_init} & 1e-6 & 1e-6 & 1e-7 & 1e-7 & 1e-7 \\
\texttt{lr\_peak} & 1e-4 & 1e-4 & 5e-5 & 3e-5 & 3e-5 \\
\texttt{lr\_end} & 1e-5 & 1e-5 & 5e-6 & 3e-6 & 3e-6 \\
\texttt{lr\_warmup\_steps} & 1.5e4 & 1.5e4 & 1.5e4 & 1.5e4 & 1.5e4 \\
% \texttt{lr\_decay\_steps} & 2.7e6 & 2.7e6 & 2.7e6 & 2.7e6 & 2.7e6 \\
\texttt{adamw\_beta1} & 0.9 & 0.9 & 0.9 & 0.9 & 0.9 \\
\texttt{adamw\_beta2} & 0.95 & 0.95 & 0.95 & 0.95 & 0.95 \\
\texttt{adamw\_weight\_decay} & 1e-3 & 1e-3 & 1e-3 & 1e-3 & 1e-3 \\
\end{tabular}
\vspace{0.75cm}
\end{table*}

%%% END: TABLE PAGE 2 %%%

\subsubsection{Action Cross-attention}
\label{sec:xattn-ablation}
In addition to using a perceiver backbone for our model, a key design choice was also to use the action as output query rather than having it as an input.
This has the advantage that we can use a common encoder for the policy and the Q-value estimator and enables quick evaluation of actions because the latent can be cashed.
Furthermore, this shortens the path from actions to Q-values and the action is thus less likely to be ignored.
In the last column of \Cref{tab:ablations} we ablate this architectural choice against using the action as input modality for control suite tasks, where we made use of attention masking in order to hide the action input from the policy output.
We observe a decrease in performance, likely due to the aforementioned reasons.

\begin{table*}[h!]
\centering
\caption{
    Different architectural ablations trained separately on Control Suite (XS-sized model) and RoboCat pyramid and tower (S-sized model).
    For Gato:Control we report the percentage of achieved expert average reward and the standard-error-based 95\% CIs.
    For RoboCat (RC) we report the average success rates and their corresponding Wilson score intervals for $\alpha_W=0.05$.
    We use 100 trials per task.
    Best results (within CI of the best mean) in each row are bold.
}
\label{tab:ablations}
\begin{tabular}{lc|ccc|c}
\textbf{Domain} & $\#(\mathcal{T})$ & \textbf{Linear Encoder} & \textbf{$\mu$-Law Encoder} & \textbf{Multi-scale Normalizer} & \textbf{w/o output action} \\
\hline
Gato:Control & 32 & 79.6~\tiny[74.5, 84.6] & \textbf{82.7~\tiny[77.2, 88.1]} & \textbf{86.7~\tiny[82.4, 90.9]} & 76.8~\tiny[72.6, 81.1] \\
\hline
\hline
RC:Tower & 7 & \textbf{68.1~\tiny[64.6, 71.5]} & 50.7~\tiny[47.0, 54.4] & \textbf{69.1~\tiny[65.6, 72.5]} & N/A \\
RC:Pyramid & 30 & 63.5~\tiny[61.7, 65.2] & 36.1~\tiny[34.4, 37.8] & \textbf{65.8~\tiny[64.1, 67.5]} & N/A \\
\end{tabular}
\end{table*}

\subsubsection{Task Conditioning}
\label{sec:dropout}

\begin{table*}[h!]
\centering
\caption{
    Adapting a pre-trained \PACQ~model (size M) to a new task-conditioning setup during finetuning phase setting $p_{\tau^V}=0.99, p_{\tau^L}=0.9$.
    The average success rates across 100 trials per task and their corresponding Wilson score intervals for $\alpha_W=0.05$ are reported.
}
\label{tab:task-conditioning-finetune}
\begin{tabular}{l|lc|cccc}
\textbf{Model} & \textbf{Domain} & $\#(\mathcal{T})$ & \textbf{Vision+Lang} & \textbf{Vision} & \textbf{Lang} & \textbf{No Goal} \\
\hline
\multirow{2}{*}{\begin{tabular}[c]{@{}l@{}}PT@3M\end{tabular}} & RC: Tower & 7 & 59.7~\tiny[49.6, 69.3] & 0.1~\tiny[0.0, 3.9] & 58.1~\tiny[48.1, 67.7] & 0.0~\tiny[0.0, 3.6] \\
& RC: Pyramid & 30 & 53.3~\tiny[43.2, 63.1] & 11.4~\tiny[10.3, 12.6] & 53.1~\tiny[43.1, 62.9] & 11.0~\tiny[9.9, 12.2] \\
\hline
\multirow{2}{*}{\begin{tabular}[c]{@{}l@{}}FT@200K\end{tabular}} & RC: Tower & 7 & 62.0~\tiny[51.8, 71.5] & 44.7~\tiny[35.4, 54.3] & 60.1~\tiny[50.1, 69.6] & 8.7~\tiny[7.1, 15.2] \\
& RC: Pyramid & 30 & 53.1~\tiny[43.1, 62.9] & 47.7~\tiny[37.7, 57.8] & 53.5~\tiny[43.5, 63.3] & 8.2~\tiny[4.3, 14.9] \\
\end{tabular}
\end{table*}

\begin{table*}[h!]
\centering
\caption{
    Pre-training a \PACQ~model (size S) with varying goal conditions.
    Evaluation after 500k training steps.
    The average success rates across 100 trials per task and their corresponding Wilson score intervals for $\alpha_W=0.05$ are reported.
}
\label{tab:task-conditioning-pretrain}
\begin{tabular}{l|lc|cccc}
\textbf{Model} & \textbf{Domain} & $\#(\mathcal{T})$ & \textbf{Vision+Lang} & \textbf{Vision} & \textbf{Lang} & \textbf{No Goal} \\
\hline
\multirow{2}{*}{\begin{tabular}[c]{@{}l@{}}$p_{\tau^V}=1.0$\\$p_{\tau^L}=1.0$\end{tabular}} & RC: Tower & 7 & 55.6~\tiny[43.4, 66.3] & 5.57~\tiny[4.0, 7.5] & 57.0~\tiny[47.0, 66.6] & 5.1~\tiny[3.6, 7.1] \\
& RC: Pyramid & 30 & 53.1~\tiny[43.0, 62.9] & 0.933~\tiny[0.7, 5.1] & 51.9~\tiny[41.9, 61.7] & 0.9~\tiny[0.9, 5.0] \\
\hline
\multirow{2}{*}{\begin{tabular}[c]{@{}l@{}}$p_{\tau^V}=0.9$\\$p_{\tau^L}=0.9$\end{tabular}} & RC: Tower & 7 & 52.6~\tiny[42.6, 62.4] & 54.6~\tiny[44.6, 64.3] & 51.4~\tiny[41.5, 61.3] & 6.3~\tiny[2.5, 12.9] \\
& RC: Pyramid & 30 & 48.2~\tiny[38.3, 58.2] & 46.8~\tiny[36.9, 56.8] & 49.8~\tiny[39.9, 59.8] & 13.1~\tiny[7.4, 21.2] \\
\end{tabular}
\end{table*}

In this ablation study we investigate how PAC uses its two different task modalities: the goal image $\tau^V$ and the language instruction $\tau^L$.
Specifically, we probe whether a trained model can react to task specifications where one of the modalities is missing and whether it can be adapted to perform better on task conditions which differ from the training setup.
To conduct this investigation, we take a \PACQ~model of size M that has been pre-trained on all data (Control Suite, CHEF and RoboCat).
Note that out of all these domains, Robocat Tower and Robocat Pyramid are the only ones that have visual task descriptions available besides language task descriptions  while all other domains have only language ones and therefore their visual task descriptions are merely padded zeros.
We then evaluate this model with different modalities for task specification being available:
\textbf{Vision+Lang}: both modalities $\tau^V$ and $\tau^L$ are present therefore the same as the setup during pretraining;
\textbf{Vision}: only visual task descriptions $\tau^V$ are present and the language ones are masked out;
\textbf{Lang}: only $\tau^L$ are available;
\textbf{No Goal}: both task description modalities are masked out.

The evaluation results are presented in the \textbf{Pretrained} rows of \Cref{tab:task-conditioning-finetune}.
A notable observation is that the pretrained model is still very performant when only the language task description $\tau^V$ is available,
albeit this differs from the task conditioning during pretraining where the model has only seen visual and language task descriptions both being present for the Tower and Pyramid domains;
while the success rate drops substaintially when only visual task descriptions $\tau^V$ are present.
One hypothesis of such an imbalanced performance of the pretrained model when conditioning on $\tau^V$ or $\tau^L$ could be the following:
Since all data domains the model has seen during the pretraining phase have $\tau^V$ available while Tower and Pyramid are the only ones with $\tau^V$,
the $N^L=50$ language task description tokens are attended to for every single datapoint while
the $N^V_{\tau}\times N_E=300$ visual task description tokens are masked out except for when learning on the Tower and Pyramid domains of robocat.
Therefore we suspect the model would learn to attend more to the language task tokens than the vision task tokens in order to be able to achieve good performance averaging over all task domains.
We conduct additional experiments for this hypothesis and present the results in \Cref{tab:task-conditioning-pretrain} which will be discussed directly following.

In order to see if above said model could be quickly adapted such that its performance is more amenable to varying modalities for task description,
we conduct a finetuning experiment of the model where either task description modality will be present for a certain percentage $p_{\tau}$ of the time and masked out the rest of the time.
In particular,
for this experiment we set $p_{\tau^V}=0.99$ and $p_{\tau^L}=0.9$ where $\tau^V$ are masked out less often to compensate for the fact that visual task descriptions are only present for two domains while the finetuning is carried over all tasks, but with adjusted sampling weights compared to~\Cref{tab:datasets-summary}: from $\lambda = (8, 2, 2, 1, 1, 2)$ to $\lambda = (6, 2, 2, 6, 6, 2)$.
We also use a fixed learning rate of 3e-6.
The task conditioning evaluation results of this finetuned model (after 200K steps) are shown in the \textbf{FT} rows of \Cref{tab:task-conditioning-finetune}.
We observe that the success rate of the model increases by a large margin when only visual task description is present compared to the pretrained model,
while the model is able to keep (or slightly improves) its performance for the conditions where it already perform well after pretraining: \textbf{Vision+Lang} and \textbf{Lang}.
This is promising as it shows the flexbility of the proposed model that it is quickly adaptable to be reactive to task specification conditions not encountered during training.

With \Cref{tab:task-conditioning-finetune} showing viable approaches to adapt pretrained models to varying task conditionings,
as just mentioned,
we conduct an extra experiment to test the hypothesis that attributes the imbalanced performance of the pretrained model under $\tau^V$ only versus $\tau^L$ only to these two modalities being improportionally present in the pretraining data.  
The experiment that tests this hypothesis is where we train a \PACQ~model (size S) from scratch, but on RoboCat Tower and RoboCat Pyramid only such that both visual and language task descriptions are available for every datapoint.
We then evaluate this model after 500k pretraining steps and present the results in the $\mathbf{p_{\tau^V}=1.0,p_{\tau^L}=1.0}$ rows of \Cref{tab:task-conditioning-pretrain}.
However, we find that with balanced presence of $\tau^V$ and $\tau^L$,
the model is still much more reactive to $\tau^L$ 
(reaching success rate of $57.0\%$ and $51.9\%$)
than $\tau^V$
($5.6\%$ and $0.9\%$).
This refutes the hypothesis and we suspect that the performance imbalance might be caused by the fact that the language task descriptions is simply more discriminative than a goal image.
% Verifying this hypothesis is however beyond the scope of this paper and can be investigated in future work.

Setting the modality present rate to $\mathbf{p_{\tau^V}=0.9,p_{\tau^L}=0.9}$,
we train another model from scratch with otherwise the same setup of above and present the evaluation results in the corresponding rows in \Cref{tab:task-conditioning-pretrain}.
The results show that with varying goal conditionings being present in the pretraining phase, the resulting model can perform well on all conditioning variations as expected.

\begin{figure*}[h!]
    \centering
    \includegraphics[width=0.99\textwidth]{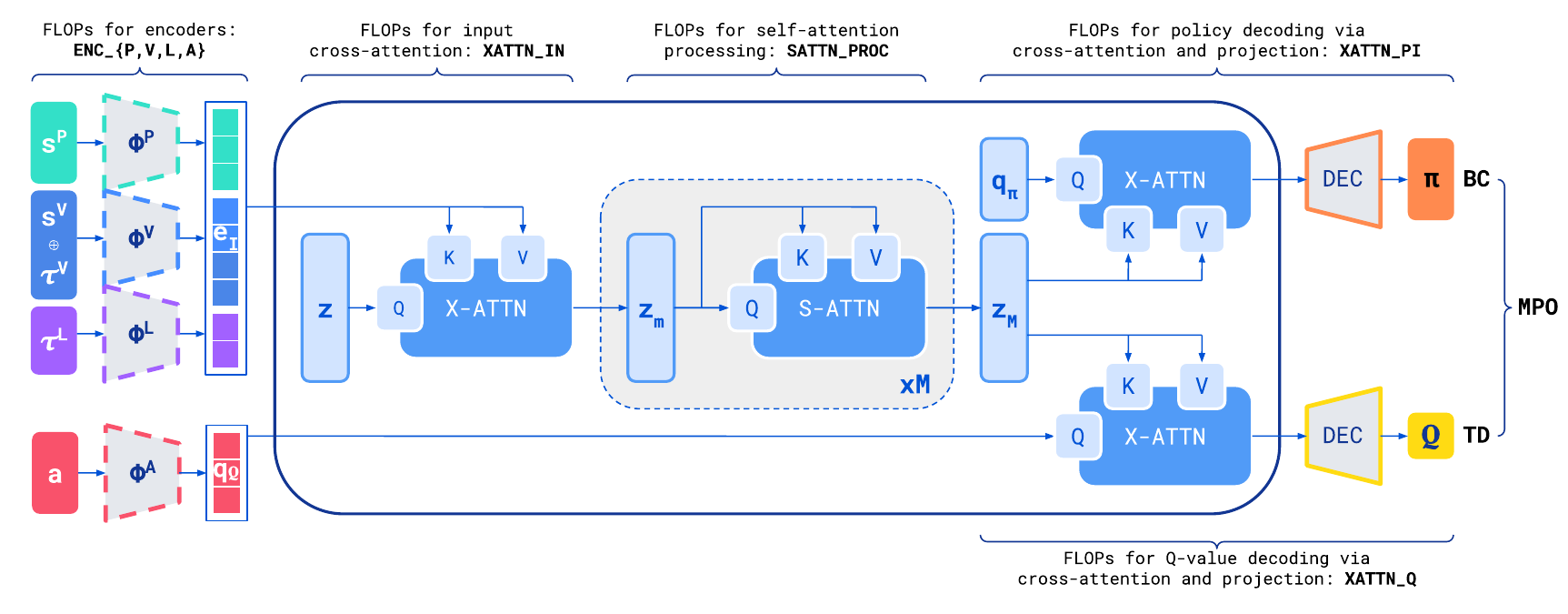}
    \caption{
    The different parts of the PAC architecture for which FLOPs are counted.
    The FLOPs for the different encoders are counted as \texttt{ENC\_\{P,V,L,A\}}, respectively.
    All FLOPs used to aggregate the encoded input tokens $e_I$ into the Perceiver's latent $z_i$ during the input cross-attention are counted as \texttt{XATTN\_IN}.
    The FLOPs used for processing the latents via $M$ self-attention blocks are counted as \texttt{SATTN\_PROC}.
    The FLOPs used to decode the policy $\pi$ and the action-value function $Q$ from $z_M$ via cross-attention and subsequent projections are counted as \texttt{XATTN\_PI} and \texttt{XATTN\_Q} respectively.
    }
    \label{fig:pac-arch-flops}
\end{figure*}

\section{Scaling Details}

\subsection{Return Profile Fitting}
\label{sec:return-profile-fitting}

We fit the average return of a models as a logistic function of the number of training steps:

\begin{equation}
    R(n) = \frac{a}{1 + \exp(-k * (n - n_0))} + b .
    \label{eq:reward-fit}
\end{equation}

We evaluate six checkpoints of each training run evenly spaced across the 3M updates.
Each evaluation runs 100 trials per task and computes the average episode return across 102 simulation tasks.
The parameters $a, k, n_0, b$ of~\Cref{eq:reward-fit} are then fitted to the evaluation data points to obtain a \emph{return profile} for each model.

\subsection{FLOP Counting}
\label{sec:flop-counting}

\begin{table}[h!]
    \footnotesize
    \centering
    \begin{tabular}{l|ccc}
         \textbf{Model Scale} & \textbf{\texttt{FWD}} & \textbf{\texttt{BWD}} & \textbf{\texttt{BATCH\_UPDATE}} \\
         \hline
         XXS (32M) & 7.826E+09 & 1.573E+10 & 1.206E+13 \\
         XS (73M) & 1.360E+10 & 2.733E+10 & 2.096E+13 \\
         S (164M) & 2.341E+10 & 4.705E+10 & 3.607E+13 \\
         M (391M) & 4.380E+10 & 8.804E+10 & 6.750E+13 \\
         L (988M) & 1.040E+11 & 2.090E+11 & 1.602E+14 \\
    \end{tabular}
    \caption{
        FLOPs costs for forward pass, backward pass and batch update for all scales of our \PACQ~model family.
        All FLOPs are computed for a batch size $B=512$ and a target update frequency $f_{\theta'}=100$.
    }
    \label{tab:pacq-flops}
\end{table}

When counting the FLOPs of the PAC architecture for the scaling analysis, we follow the FLOP counting scheme established by~\citet{hoffmann2022chinchilla} since the bulk of our model's computation sits in cross-attention and self-attention blocks.
FLOPs of cross-attention blocks are counted similarly as in self-attention blocks with the only difference of the length of the input and output sequence being different which yields \texttt{seq\_len\_in} $\times$ \texttt{seq\_len\_out} $\times$ \texttt{FLOPS\_ATTN} instead of \texttt{seq\_len\_in} $\times$ \texttt{seq\_len\_in} $\times$ \texttt{FLOPS\_ATTN} like in a normal self-attention block.

After tokenization of proprioception, vision, language and action  (cf.~\Cref{sec:tokenization}) we obtain $T^P$, $T^V$, $T^A$ and $T^L$ tokens respectively.
Their encoders are simple projections and we count the FLOPs used for the embeddings as:
\begin{itemize}
    \item $\texttt{ENC\_P} = \texttt{MAF} \times T^P \times D_I$
    \item $\texttt{ENC\_V} = \texttt{MAF} \times T^V \times D_I + \texttt{FLOPS\_RESNET}$
    \item $\texttt{ENC\_L} = \texttt{MAF} \times T^L \times N_T \times D_I$
    \item $\texttt{ENC\_A} = \texttt{MAF} \times T^A \times D_I$
\end{itemize}
Similar to~\citet{hoffmann2022chinchilla} we use a \emph{multiply-accumulate-factor} (\texttt{MAF}) of 2 to describe the multiply accumulate cost.
The FLOPs needed to transform the raw input images into tokens using a ResNet are captured by \texttt{FLOPS\_RESNET}.
We count each 2D convolution operation in the ResNet as $\texttt{num\_kernels} \times (w_1 * w_2) \times (o_1, o_2) \times \texttt{MAF}$ where $w_{\{1,2\}}$ and $o_{\{1,2\}}$ are the kernel and output dimensions respectively.

The total number of FLOPs used for one forward pass of \PACQ~are:
\begin{align*}
    \texttt{FLOPS\_FWD} & = \texttt{ENC\_P} + \texttt{ENC\_V} + \texttt{ENC\_L} + \texttt{ENC\_A} \\
    & + \texttt{XATTN\_IN} \\
    & + M \times \texttt{SATTN\_PROC} \\
    & + \texttt{XATTN\_PI} + \texttt{XATTN\_Q}
\end{align*}
When estimating the FLOPs for the backward pass, we slightly deviate from~\citet{kaplan2020scaling} and also factor in the target network updates (cf~\Cref{sec:method-alg}) which occur every $f_{\theta'}$ updates (in all of our experiments we keep $f_{\theta'}=100$).
Therefore, we count the FLOPs for PAC's backward pass as:
\begin{equation*}
    \texttt{FLOPS\_BWD} = (2 + \frac{1}{f_{\theta'}}) * \texttt{FLOPS\_BWD}
\end{equation*}
Lastly, the FLOPs for an update with batch size $B$ are counted as:
\begin{equation*}
    \texttt{FLOPS\_UPDATE} =  B * (\texttt{FLOPS\_BWD} + \texttt{FLOPS\_BWD})
\end{equation*}
We list the FLOPs costs for the core operations of our \PACQ~model series in~\Cref{tab:pacq-flops}.

\begin{figure}[h!]
    \begin{tabular}{lc|cc}
    \multicolumn{4}{c}{\includegraphics[width=0.95\columnwidth]{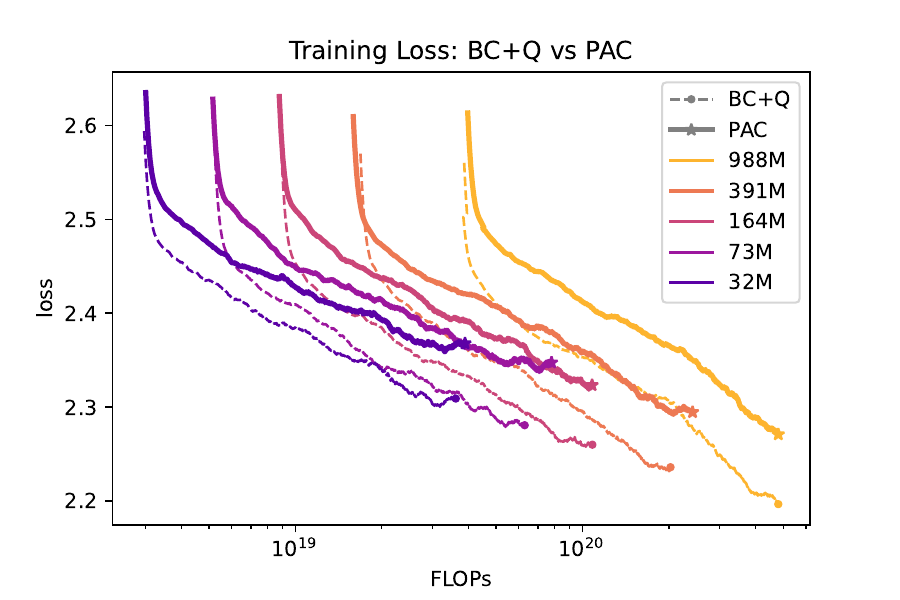}} \\
    \textbf{Domain} & $\#(\mathcal{T})$ & \textbf{BC+Q} & \textbf{PAC} \\
    \hline
    Gato: Control & 32 & 89.2~\tiny[84.3, 94.1] & 93.2~\tiny[89.6, 96.9] \\
    RC: Tower & 7 & 62.9~\tiny[59.1, 66.2] & 63.1~\tiny[59.5, 66.6] \\
    RC: Pyramid & 30 & 50.5~\tiny[48.7, 52.3] & 55.6~\tiny[53.8, 57.4] \\
    RC: Insertion & 3 & 82.0~\tiny[76.9, 85.6] & 83.7~\tiny[79.1, 87.4] \\
    CHEF: sim & 1 & 12.0~\tiny[7.0, 19.8] & 40.0~\tiny[30.9, 49.8] \\
    \end{tabular}
    \caption{
    Top: Training loss comparison of the PAC model families for $\alpha=1.0$ (\BC) and $\alpha=0.75$ (\PACQ).
    Bottom: Success rates in different simulation tasks of the final checkpoints after 3M updates for the respective L-size models.
    Each model was evaluated in 100 trials on each of the $\#(\mathcal{T})$ in each domain.
    For the Control domain, the percentage of achieved expert average reward and the standard-error-based 95\% confidence intervals are reported.
    For all other domains, the average success rates and their corresponding Wilson score intervals for $\alpha_W=0.05$ are reported.
    }
    \label{fig:qpi_loss_reward}
\end{figure}

\begin{figure}[h!]
    \begin{tabular}{c}
        \includegraphics[width=0.95\columnwidth]{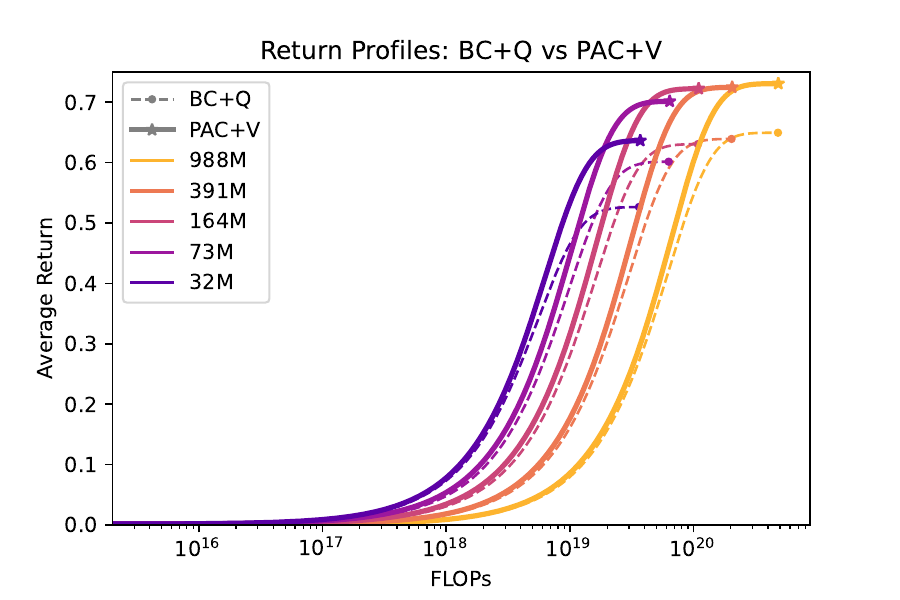} \\
        \includegraphics[width=0.95\columnwidth]{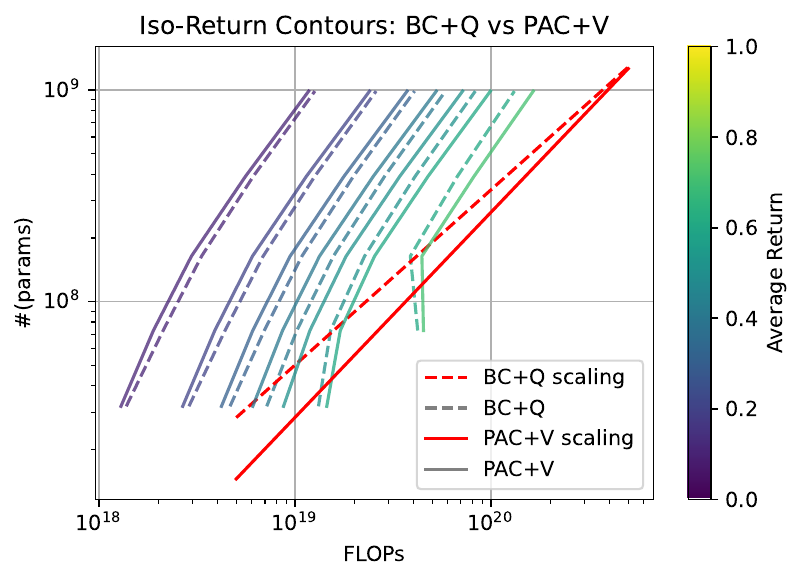} \\
    \end{tabular}
    \caption{
    Iso-Return comparison of \BC~vs \PACV.
    The return profiles (top) contrast the expected average return between the BC baseline and the RL objective across all model scales.
    The Iso-Return contours (bottom) depict how the reward landscape over the parameter-FLOPs landscape shifts between using the BC objective (dashed contours) and the RL objectives (solid contours).
    For \PACV~the reward landscape is shifted towards the top left compared to the BC baseline indicating that it leads to higher average return plateaus for the same FLOP budgets.
    }
    \label{fig:iso_reward_cmp_vpi}
\end{figure}

\begin{figure*}[ht!]
    \begin{tabular}{c}
        \includegraphics[width=0.35\textwidth]{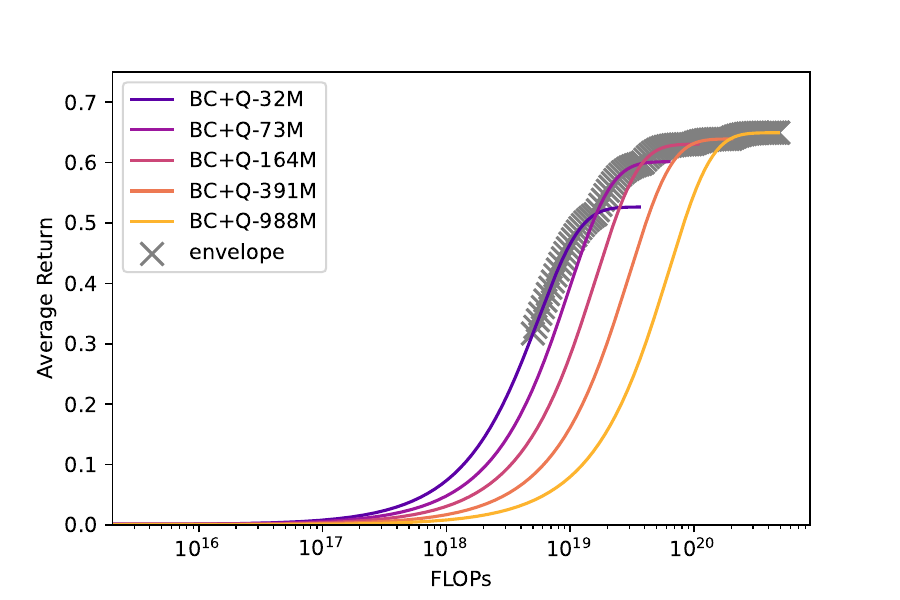}
        \includegraphics[width=0.31\textwidth]{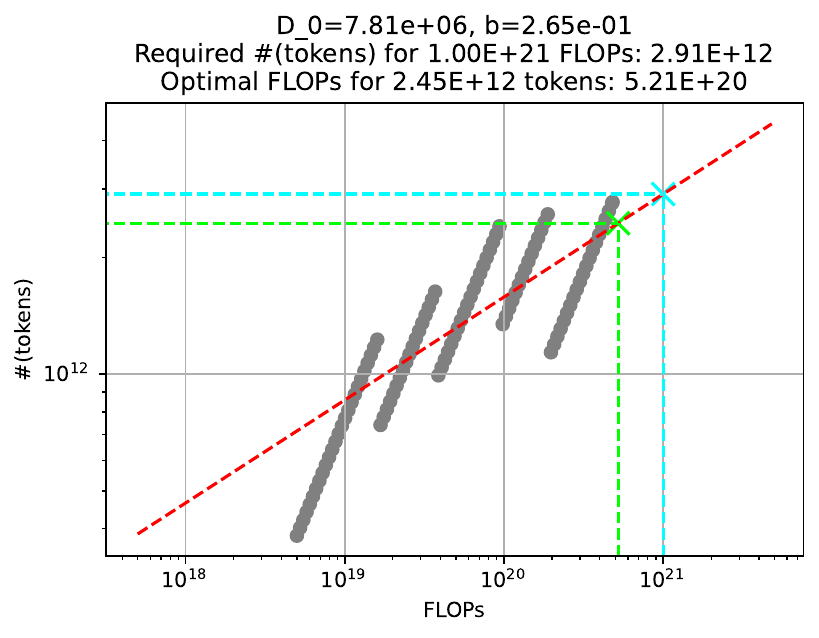}
        \includegraphics[width=0.31\textwidth]{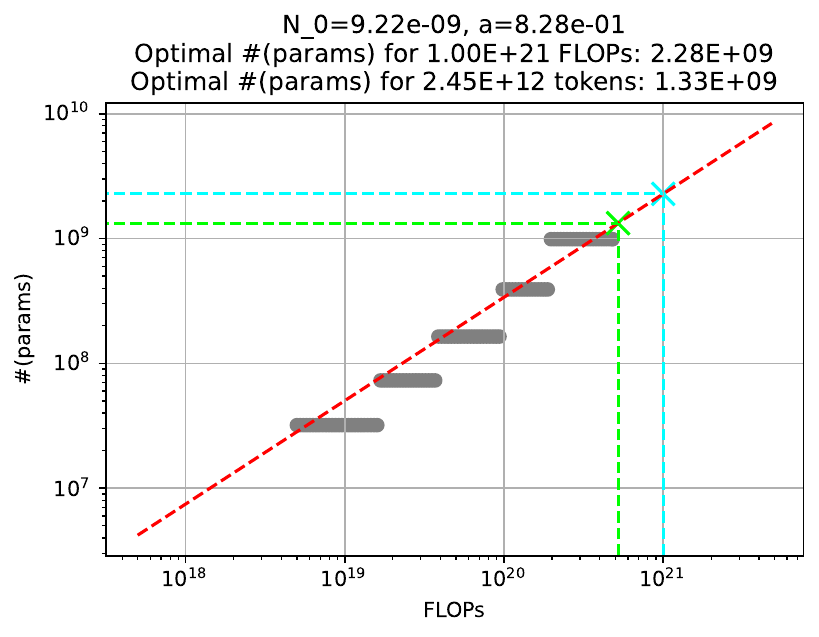} \\
        \includegraphics[width=0.35\textwidth]{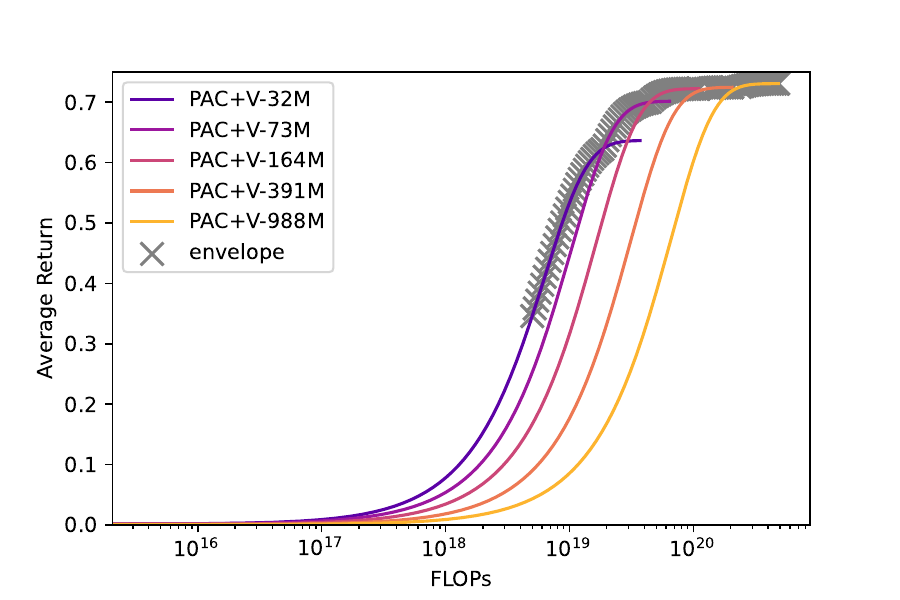}
        \includegraphics[width=0.31\textwidth]{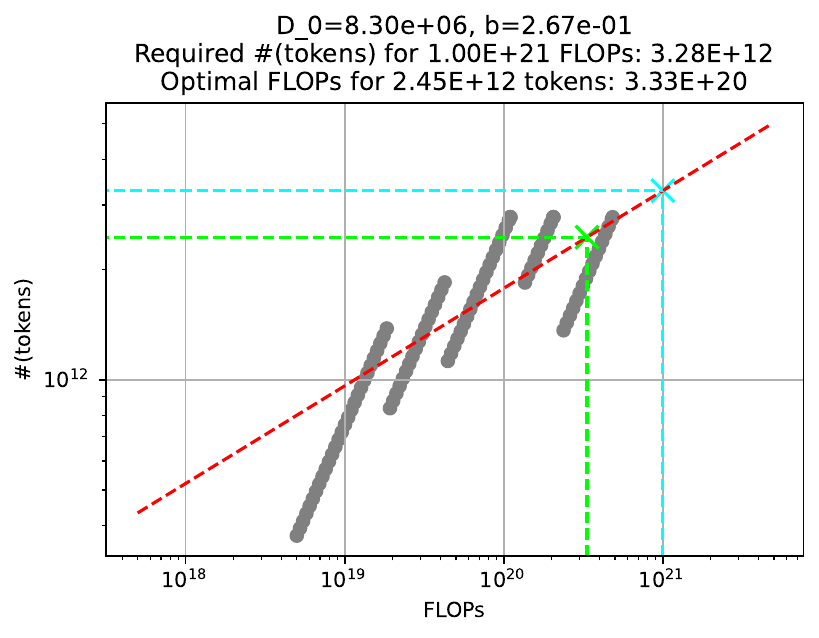}
        \includegraphics[width=0.31\textwidth]{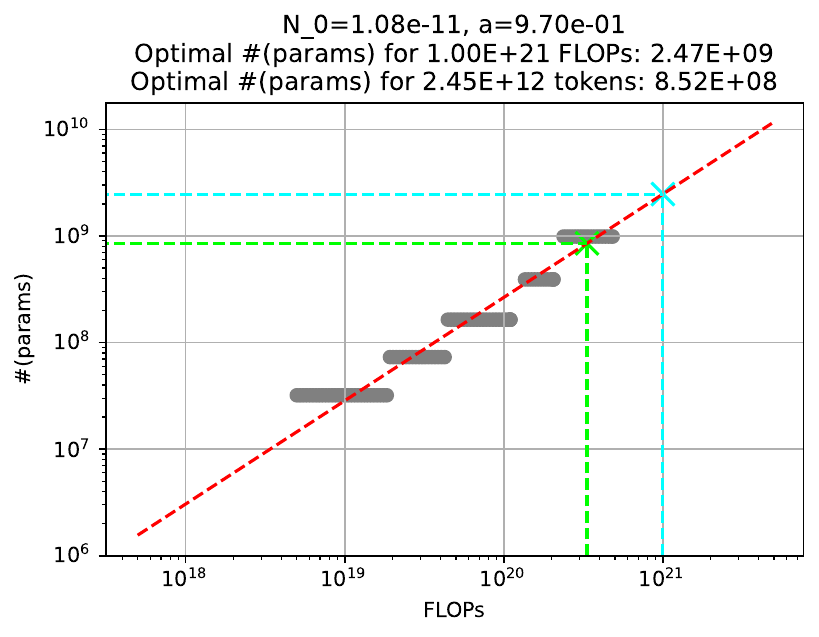} \\
    \end{tabular}
    \caption{
    Scaling laws based on the return profile envelopes for \BC~(top) and \PACV~(bottom).
    We select 100 logarithmically spaced points between 5E+18 and 5E+20 FLOPs on the envelope of the return profiles for the scaling law fits.
    For both the token and parameter scaling plots, we indicate the scaling trend with a dashed red line.
    The green intersection represents the optimality point when training on a single epoch of our data while the teal intersection represents the optimal data and parameter trade-off for a FLOP budget of 1E+21.
    }
    \label{fig:reward_scaling_ext}
\end{figure*}

\subsection{Model Loss as Performance Indicator}
\label{sec:loss-scaling}

We plot the training loss against the FLOPs for both model sets in~\Cref{fig:qpi_loss_reward} when training on the scaling data mix (cf.~\Cref{tab:data-mixture}).
The training losses suggest that the \BC~model should outperform \PACQ~because its final loss is about 0.1 points lower.
However, as in previous works~\citep{hilton2023scaling} we find that loss is not necessarily a reliable indicator of model performance in RL settings.
In order to assert whether the training loss can be a reliable indicator of model performance in our experiments, we compare the final model checkpoints of \BC and \PACQ~(both of size L) on 73 tasks from our training dataset in simulation and report the results in~\Cref{fig:qpi_loss_reward}.
However, this performance-based comparison does not support the assumption that a lower training loss is indicative of higher task performance.
On the contrary, PAC significantly outperforms BC on the simulated stacking task despite its higher training loss.
Hence, we cannot use the model directly for selecting interpolants to fit the scaling laws with and need use the proxy of return profiles (cf.~\Cref{sec:return-profile-fitting}).

\subsection{\PACV~Scaling Results}
\label{sec:scaling-ext}

We also conduct a scaling analysis for the V-function variant of our architecture: \PACV~(cf.~\Cref{sec:pac-v-app}) following the protocol defined in~\Cref{sec:scaling}.
We plot the scaling laws for \PACV~in~\Cref{fig:reward_scaling_ext} comparing it to the scaling laws for the \BC~baseline as well.
We also compare the Iso-Return contours between the two models in~\Cref{fig:iso_reward_cmp_vpi}.

The conclusions of the \PACV~scaling laws are consistent with the ones drawn for the Q-function variant of \PACQ~in \Cref{sec:scaling}.
The suggested model size for 2.45T tokens is with 852M parameters slightly smaller than \PACQ's with 954M parameters, but the parameters should be scaled slightly more aggressively as compute increases indicated by $a(\PACV) > a(\PACQ)$: $0.970 > 0.920$.
However, the data scaling is in line with both \PACQ~as well as the \BC~baseline: $b(\PACV) \approx 0.266$.

\section{Additional Experiments}
\label{sec:additional-exp-app}

\subsection{Large-scale Training of \PACV}
\label{sec:pacv-pretraining}

As mentioned in~\Cref{sec:exp-pretraining}, we also conduct a large-scale training on the full pre-training corpus (cf.~\Cref{tab:data-mixture-pt}) following the same protocol as for the other models.
We report the additional results obtained with \PACV~in \Cref{tab:pretraining-results-ext}.

The results for the V-function variant of PAC are in line with the ones obtained using the Q-function variant with two notable exceptions.
First, in the case of RC:Insertion which consists exclusively of human teleoperated data, \PACV~provides another significant improvement over \PACQ~to$\approx 89\%$ success rate.
This suggests that \PACV~could have an edge in cases where many successful task demonstrations exist which are however beset by minor inefficiencies, e.g. motion jitter or pauses caused by human teleoperation.
The results suggest that the one-step improvement over the data-generating policy afforded by \PACV~could be enough to prevent imitating many inefficiencies present in the training data.
Second, in the case of CHEF:sim, \PACV~lags behind its Q-function-based counterpart \PACQ.
This suggests that in cases where the data is mostly sub-optimal, a value function alone might not be sufficient to filter transitions effectively enough to improve the policy.

\begin{table*}[ht!]
\centering
\caption{
Policy success rates across $\#(\mathcal{T})$ tasks in each domain for 100 evaluations per task.
The average success rate in the training data is reported as $p_D$.
For Gato: Control, the percentage of achieved expert average reward and the standard-error-based 95\% CIs are reported (where available).
For all other task families, the average success rates and their corresponding Wilson score intervals for $\alpha_W=0.05$ are reported.
Best results (within CI of the best mean) in each row are bold.
\scriptsize[$\dagger$ cited from \citet{reed2022generalist}; $\filledstar$ cited from \citet{bousmalis2023robocat}]
}
\label{tab:pretraining-results-ext}
\begin{tabular}{lcc|ccc|cc}
\textbf{Domain} & $\#(\mathcal{T})$ & $p_D$ & \textbf{Gato$\dagger$ / RC$\filledstar$} & \textbf{FilteredBC} & \textbf{BC+Q} & \textbf{PAC} & \textbf{PAC+V} \\
\hline
Gato:Control & 32 & N/A & 63.6$\dagger$ & 75.8~\tiny[62.5, 78.6] & 84.6~\tiny[79.6, 89.7] & \textbf{87.7~\tiny[83.8, 91.6]} & \textbf{90.2~\tiny[86.3, 94.1]} \\
\hline\hline
RC:Tower & 7 & 75 & 61.0$\filledstar$~\tiny[57.3, 64.5] & 64.0~\tiny[60.4, 67.5] & \textbf{71.3~\tiny[67.8, 74.5]} & \textbf{69.3~\tiny[65.8, 72.6]} & 67.3~\tiny[63.7, 70.7] \\
RC:Pyramid & 30 & 75 & \textbf{64.5$\filledstar$~\tiny[62.8, 66.2]} & \textbf{64.0~\tiny[62.3, 65.7]} & 62.4~\tiny[60.7, 64.1] & \textbf{63.5~\tiny[61.7, 65.1]} & \textbf{65.3~\tiny[63.5, 66.9]} \\
RC:Insertion & 3 & 97 & 71.3$\filledstar$~\tiny[66.0, 76.2] & 81.0~\tiny[75.8, 84.7] & 79.7~\tiny[74.8, 83.8] & 80.3~\tiny[75.5, 84.4] & \textbf{89.0~\tiny[85.0, 92.1]} \\
CHEF:sim & 1 & 28 & N/A & 17.0~\tiny[10.9, 25.5] & 11.0~\tiny[6.3, 18.6] & \textbf{55.0~\tiny[45.2, 64.4]} & 42.0~\tiny[32.8, 51.8] \\
\end{tabular}
\end{table*}

\begin{table*}[ht!]
\centering
\caption{
Per-group success rates and confidence intervals for real-robot stacking tasks across self-improvement iterations ($\#(\mathcal{T})=5$).
The policies are evaluated in 400 trials per set.
The average success rates and their corresponding Wilson score intervals for $\alpha_W=0.05$ are reported.
}
\begin{tabular}{l|ccccc|c}
\textbf{Iteration} & \textbf{Set 1} & \textbf{Set 2} & \textbf{Set 3} & \textbf{Set 4} & \textbf{Set 5} & \textbf{All Sets} \\ \hline
\textbf{Pretraining} & 51.5 \tiny{[46.6, 56.4]} & 53.5 \tiny{[48.6, 58.3]} & 65.5 \tiny{[60.7, 70.0]} & 84.7 \tiny{[80.8, 87.9]} & 94.0 \tiny{[91.2, 95.9]} & 69.8 \tiny{[67.8, 71.8]} \\
\textbf{RLFT \#1} & 83.0 \tiny{[79.0, 86.4]} & 66.8 \tiny{[62.0, 71.2]} & 87.5 \tiny{[83.9, 90.4]} & 90.8 \tiny{[87.6, 93.3]} & 95.5 \tiny{[93.0, 97.1]} & 84.7 \tiny{[83.1, 86.2]} \\
\textbf{RLFT \#2} & 88.0 \tiny{[84.4, 90.8]} & 76.2 \tiny{[71.8, 80.1]} & 92.5 \tiny{[89.5, 94.7]} & 95.8 \tiny{[93.4, 97.4]} & 96.8 \tiny{[94.6, 98.1]} & 89.8 \tiny{[88.4, 91.1]} \\
\textbf{RLFT \#3} & 91.8 \tiny{[88.7, 94.1]} & 91.5 \tiny{[88.4, 93.9]} & 90.8 \tiny{[87.6, 93.3]} & 95.0 \tiny{[92.4, 96.7]} & 97.0 \tiny{[94.8, 98.3]} & 93.2 \tiny{[92.0, 94.2]} \\
\end{tabular}
\label{tab:stacking-detail-results}
\end{table*}

\subsection{RL Fine-tuning and Self-improvement}
\label{sec:stacking-detail-results}

In \Cref{sec:exp-rlft} we evaluated \PACQ on a robotic stacking benchmark, and performed iterative fine-tuning to improve performance in this domain.
Here, we provide additional details for those results.

The RGB Stacking benchmark~\citep{lee2021beyond} defines five distinct sets of test objects, each highlighting a different challenge of object manipulation.
For brevity, we only reported mean success rates above.
In \Cref{tab:stacking-detail-results}, we provide success rates for each object separately.
The continuous self-improvement of \PACQ is particularly visible on "Set 2", which requires precise force-based control to flip objects onto their side.
However, the same holds for the other object sets, which improve across fine-tuning rounds until converging at $>90\%$ success rate.

Data collection was carried out only for the CHEF:real domain, so it is worth examining whether such focused self-improvement causes the PAC model's performance on other tasks to degrade.
As \Cref{tab:rlft-results-ext} shows, performance on the simulated tasks is unaffected, even after three rounds of fine-tuning.

\begin{table}[h!]
\tabcolsep=0.12cm
\centering
\caption{
Comparison of performance across all domains after pre-training (\aPAC), after three rounds of self-improvement on the CHEF:real domain (RLFT \#3).
}
\begin{tabular}{lc|ccc}
    \textbf{Domain} & $\#(\mathcal{T})$ & \textbf{$\alpha$-PAC} & \textbf{RLFT \#3} \\%& \textbf{RLFT \#4} \\
    \hline
    Gato: Control & 32 & 92.1 & 91.3  \\%& 94.3\\
    ~ & ~ & \tiny[88.4, 95.9] & \tiny[89.6, 96.9] \\%& \tiny[91.3, 97.3] \\
    RC: Tower & 7 & 69.6 & 70.0 \\%& 70.3 \\
    ~ & ~ & \tiny[65.9, 72.7] & \tiny[66.5, 73.3] \\%& \tiny[66.8, 73.6] \\
    RC: Pyramid & 30 & 64.9 & 65.1 \\%& 65.7 \\
    ~ & ~ & \tiny[63.1, 66.6] & \tiny[63.3, 66.8] \\%& \tiny[64.0, 67.4] \\
    RC: Insertion & 3 & 89.3 & 80.3 \\%& 83.0 \\
    ~ & ~ & \tiny[85.0, 92.1] & \tiny[75.5, 84.4] \\%& \tiny[78.3, 86.8] \\
    CHEF: sim & 1 & 52.0 & 59.0 \\%& 51.0 \\
    ~ & ~ & \tiny[42.3, 61.5] & \tiny[49.2, 68.1] \\%& \tiny[41.3, 60.6] \\
    CHEF: real & 5 & 69.8 & 93.2 \\%& 86.2 \\
    ~ & ~ & \tiny[67.8,71.8] & \tiny[92.0,94.2] \\%& \tiny[84.6, 87.6] \\
    \end{tabular}
    \label{tab:rlft-results-ext}
\end{table}

While we started the iterative improvement in \Cref{sec:exp-rlft} by pre-training \aPAC, the BC/RL trade-off parameter $\alpha$ also allows flexibility as to what data to start from.
For comparison, we also deploy \BC on the robot -- a model which is pre-trained using only the BC part of the objective to optimize its policy ($\alpha=1$), but which has already learned a Q-function on the pre-training data ($\beta>0$).
The initial performance of this model on the real robot is unsurprisingly low with only $3.6\%$ success rate.
When we continue the training on the \emph{same pre-training data} (including all sim and real tasks) for another 200k updates, but lower $\alpha$ to $0$ to fully leverage the Q-function for policy improvement, we observe a significant jump to $38.2\%$ success rate when re-deploying on the robot.
While this performance is still significantly lower than the $69.8\%$ success of the \aPAC model after initial pre-training, it is high enough to feasibly be used as an alternate starting point for the self-improvement loop.
This can be useful in scenarios where no non-expert data is available initially to perform \aPAC from the start.

\end{document}